\documentclass{article}

\usepackage[numbers]{natbib}
\usepackage[eandd, final]{neurips_2026}

\usepackage[utf8]{inputenc} % allow utf-8 input
\usepackage[T1]{fontenc}    % use 8-bit T1 fonts
\usepackage{hyperref}       % hyperlinks
\usepackage{url}            % simple URL typesetting
\usepackage{booktabs}       % professional-quality tables
\usepackage{amsfonts}       % blackboard math symbols
\usepackage{nicefrac}       % compact symbols for 1/2, etc.
\usepackage{microtype}      % microtypography
\usepackage{xcolor}         % colors
\usepackage{graphicx}
\usepackage{amsmath}
\usepackage{caption} 
\usepackage{multirow}
\usepackage{tcolorbox}
\usepackage{tabularx}
\usepackage{booktabs}
\usepackage{wrapfig}
\usepackage{makecell}
\usepackage[table]{xcolor}
\usepackage{threeparttable} % \begin{tablenotes}
\usepackage{algorithm}
\usepackage{algpseudocode}
\usepackage{subcaption}

\definecolor{mygreen}{RGB}{0, 150, 0} 
\definecolor{myred}{RGB}{200, 0, 0}
\definecolor{mygray}{rgb}{0.95,0.95,0.95} % 定义浅灰色
\definecolor{darkgreen}{RGB}{45, 130, 117}
\definecolor{darkorange}{RGB}{187, 114, 14}
\definecolor{myblue}{RGB}{65, 85, 198}
\definecolor{darkred}{RGB}{192, 0, 0}
\newcommand{\myuparrow}{\textcolor{mygreen}{$\uparrow$}} 
\newcommand{\mydownarrow}{\textcolor{myred}{$\downarrow$}}

\usepackage{xcolor}

\tcbuselibrary{breakable}
\usepackage{pifont}

\title{GUITAR: Structured Failure Diagnosis of GUI Agents via State Transitions}
\author{%
    Shaoqing Zhang$^{1,2}$, 
    Xuefeng Bai$^{1}$\thanks{Corresponding authors: Xuefeng Bai and Yang Xiang.}, 
    Kehai Chen$^{1,2}$,
    Zhuosheng Zhang$^{3}$
    \and
    \textbf{Pengfei Zhang}$^{2}$,
    \textbf{Yang Xiang}$^{2}$\footnotemark[1],
    \textbf{Min Zhang}$^{1,2}$ \\
    $^{1}$Harbin Institute of Technology, Shenzhen, China \\
    $^{2}$Pengcheng Laboratory, Shenzhen, China \\
    $^{3}$Shanghai Jiao Tong University, China \\
}

\begin{document}

\maketitle

% !TeX root = ../neurips_2026.tex
\begin{abstract}

Understanding where and why Graphical User Interface (GUI) agents fail is essential for building more reliable systems, yet current evaluation relies on step-level success rate (SR), a metric that treats each screen independently and overlooks the underlying structure of GUI environments.
This leads to two critical blind spots: (1) functionally equivalent screens are evaluated in isolation, obscuring systematic failure patterns across shared screens; and (2) the long-tailed GUI distribution renders failures on rare but critical screens invisible under standard metrics.
To address these issues, we propose \textbf{GUITAR}, a state-centric diagnostic framework that maps visually diverse screens to shared functional states and uses the resulting State Transition Graph (STG) to perform structured failure analysis over both states and transitions.
Across 8 agents and 6 tasks from AndroidControl and Mind2Web, GUITAR reveals that 60.4\% of failures occur in 20\% of states, localizing errors to a small set of bottlenecks.
Bottleneck-targeted guidance improves SR by 2.8 percentage points and retains an average gain of 1.88 percentage points across 7 agents under three-fold trajectory-held-out evaluation with fully automatic STGs.
These findings demonstrate the diagnostic and actionable value of structure-aware evaluation within the evaluated mobile and web tasks.
Code is available at \url{https://github.com/sqzhang-lazy/GUITAR}.

\end{abstract}

% !TeX root = ../neurips_2026.tex
\section{Introduction}

Recent advances in Vision-Language Models (VLMs) have enabled GUI agents capable of autonomously operating complex applications~\citep{hong2024cogagent, yang2023auto, yan2023gpt, zhang2024dynamic,tang2025survey}. While current evaluation predominantly relies on step-level success rate (SR)~\citep{zhang2024you, rawles2024androidworld, zheng2025naturegaia,ji2026thinking,deng2025retail}, this metric treats each screen as an independent observation, fundamentally ignoring the underlying structure of GUI environments.

\begin{figure}[t]
    \centering
    \includegraphics[width=0.93\linewidth]{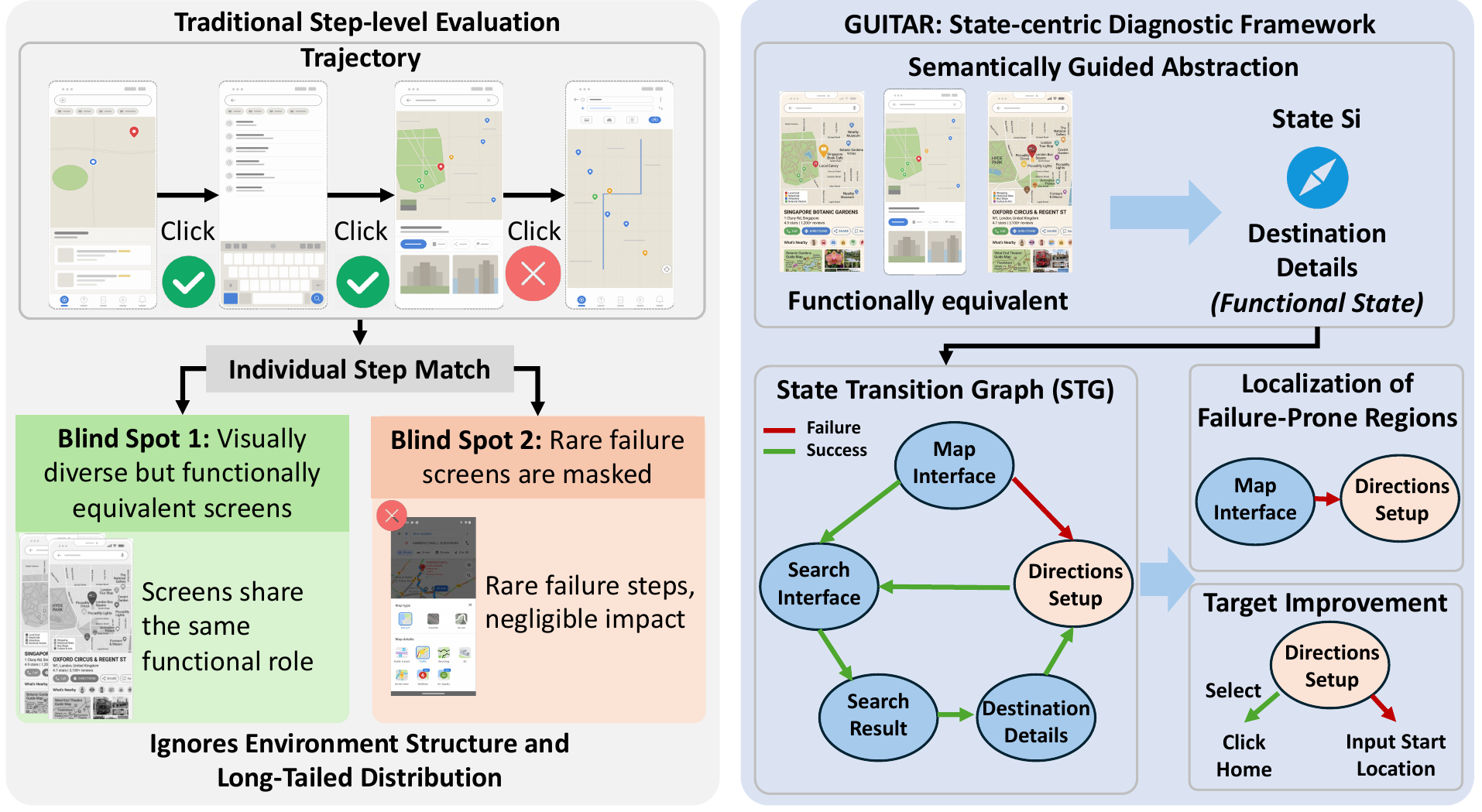}
    \caption{Limitations of step-level evaluation and the need for structured failure diagnosis. Left: step-level success rate aggregates performance across all steps, obscuring where failures occur. Right: GUITAR maps execution outcomes onto the STG, enabling state-level and transition-level analysis to localize bottleneck states and failure transitions for targeted diagnosis and improvement.}
    \label{fig:GUITAR}
    % \vspace{-3ex}
\end{figure}

This step-level view leads to two critical blind spots, as illustrated in Figure~\ref{fig:GUITAR}. 
First, many screens share the same functional role in task completion despite surface-level differences. 
Without mapping such screens to shared functional states, evaluation remains tied to raw observations rather than interaction semantics. 
This obscures systematic failure patterns across functionally equivalent states.  
Second, GUI trajectories follow a long-tailed state distribution: high-frequency states dominate SR, causing failures on rare but critical states to contribute negligibly to the overall score. 
Such failures remain effectively invisible under standard metrics. 
% Together, these limitations prevent step-level evaluation from localizing failure-prone regions, making it fundamentally inadequate for diagnosing agent behavior.

To address this challenge, we propose \textbf{GUITAR}, a state-centric evaluation framework that diagnoses GUI agent failures through structured analysis of State Transition Graphs (STGs). 
Unlike similarity-based approaches, GUITAR maps screens to functional states based on behavioral equivalence, enabling global structured failure analysis over both states and transitions.

% However, constructing such abstractions is non-trivial: GUI environments lack explicit state identities, and high visual variability makes it difficult to reliably group screens into meaningful functional states. 
% For addressing this question, we propose \textbf{GUITAR}. This state-centric diagnostic framework performs structured failure analysis over both states and transitions, using a State Transition Graph (STG) that represents agent behavior as functional state progressions rather than raw screen sequences.
% To the best of our knowledge, GUITAR is the first framework to perform global, state-level diagnosis of GUI agents.

Across 8 agents and 6 tasks from AndroidControl and Mind2Web, GUITAR reveals that failures concentrate in a small set of \textbf{bottleneck states}: 20\% of states account for 60.4\% of failures, a concentration obscured by aggregate evaluation metrics.
% Second, GUITAR's diagnostic conclusions are robust: the underlying state representations preserve behavioral consistency significantly better than similarity-based alternatives, validating the reliability of our diagnostic outputs. 
Furthermore, bottleneck-targeted guidance improves SR by 2.8 percentage points and retains an average gain of 1.88 percentage points across 7 agents under three-fold trajectory-held-out evaluation with fully automatic STGs.
We further validate that GUITAR's functional state abstraction achieves high behavioral consistency, outperforming similarity-based baselines and attaining strong inter-annotator agreement.
The main contributions of this work are as follows: 

(1) We propose \textbf{GUITAR}, a state-centric diagnostic framework that reformulates GUI agent evaluation as a graph-structured problem. By abstracting raw screens into functional states and modeling their transitions as a State Transition Graph (STG), GUITAR introduces state- and transition-level metrics that capture structural failure patterns fundamentally inaccessible to step-level evaluation.

(2) We conduct the first global, cross-trajectory analysis of GUI agent failures, revealing two previously underexplored phenomena: (i) failures on low-frequency but critical states are systematically underrepresented by standard metrics, and (ii) errors are highly concentrated on a small set of bottleneck states and transitions, exposing a strongly skewed and structured failure distribution.

(3) We demonstrate that the failures identified by GUITAR are actionable. Through a closed-loop experiment, we show that targeting bottleneck states leads to positive average gains, establishing a direct link between structured diagnosis and effective agent optimization.
% : targeting bottleneck states yields positive average gains, including under trajectory-held-out evaluation with fully automatic STGs.

% !TeX root = ../neurips_2026.tex
\section{Related Work}

\subsection{Evaluation of GUI Agents}

GUI agent evaluation has evolved from static UI understanding~\citep{bai2021uibert,deka2017rico} and element grounding~\citep{cheng2024seeclick,rawles2023androidinthewild,li2025screenspot} to trajectory-based benchmarks that evaluate agents through aggregate outcome metrics such as task success rate~\citep{zhou2024webarena,xie2024osworld,lu2024gui}. However, these benchmarks remain outcome-focused, reporting aggregate metrics without exposing where and why agents fail. 
% Recent efforts introduce process-level signals to improve diagnostic observability~\citep{gao2026guitester,zhai2026guide,yang2026probench}.
Recent efforts introduce process-level signals to improve diagnostic observability. \citet{gao2026guitester} propose exploratory defect discovery with probing and reflection mechanisms; \citet{zhai2026guide} decompose trajectories into sub-tasks for interpretable diagnosis; \citet{yang2026probench} incorporate intermediate execution correctness as a process-level supervision signal.

Despite finer granularity, failures remain bound to individual trajectory segments rather than \textbf{functional states} shared across trajectories, precluding the cross-trajectory aggregation needed to identify failure-prone states systematically.

% Recent efforts introduce process-level signals to improve diagnostic observability. \citet{gao2026guitester} proposes exploratory defect discovery with probing and reflection mechanisms; \citet{zhai2026guide} decomposes trajectories into sub-tasks for interpretable diagnosis; \citet{yang2026probench} incorporates intermediate execution correctness as a process-level supervision signal.

% Despite finer granularity, failures remain properties of trajectory segments rather than \textbf{functional states} that recur across trajectories. 
% This precludes cross-trajectory aggregation that reveals which states are systematically failure-prone, a capability absent from trajectory-level representations.

\subsection{Graph-Based Modeling of GUI Systems}

\begin{table}[t]
\centering
\caption{Comparison of GUITAR with related graph-based GUI approaches. \textbf{Cross-Task Aggregation}: cross-run trajectory aggregation into a unified graph. 
\textbf{Abstraction Level}: semantic depth of state unification, from perceptual similarity to functional equivalence. 
\textbf{Failure Localization}: whether structured failure diagnoses are produced.
% \textbf{Inference Guidance}: whether the graph supports inference-time agent guidance.
}
\label{tab:related_work_comparison}
% \centering\small
% \renewcommand{\arraystretch}{1}
% \setlength{\tabcolsep}{0.9mm}
\resizebox{\textwidth}{!}{
    \begin{tabular}{lcccc}
    \toprule
    \textbf{Method} & 
    \textbf{\makecell[c]{Cross-Task Aggregation}} & 
    \textbf{Abstraction Level} & 
    \textbf{Primary Purpose} & 
    \textbf{\makecell[c]{Failure Localization}} 
    \\
    \midrule
    Step-level SR~\citep{rawles2024androidworld} 
        & \ding{55} & N/A            & Performance Measurement  & \ding{55}  \\
    PageAgent~\citep{chen2025pg}           
        & \ding{51} & Perceptual     & Navigation Planning      & \ding{55}  \\
    WebGraphEval~\citep{qian2025webgrapheval}     
        & \ding{55} & Behavioral     & Cross-Agent Comparison   & \ding{55}   \\
    \midrule
    \textbf{GUITAR (Ours)}                       
        & \ding{51} & \textbf{Semantic-Functional} & \textbf{Failure Diagnosis} 
        & \ding{51}  \\
    \bottomrule
    \end{tabular}%
}
\end{table}

To overcome the representational limitations of trajectory-level evaluation, GUITAR draws on graph-based modeling to provide a structured substrate for failure diagnosis.
Graph representations have been widely used to model GUI structure and interaction dynamics. Early work models applications as Event Flow Graphs~\citep{memon2007event} and extracts interface structure via systematic exploration~\citep{memon2003gui,su2017guided}. 
Later approaches explore higher-level representations: \citet{li2021screen2vec} encode screens into semantic vectors.
Most relevant to our work are approaches that construct graph structures over GUI trajectories for evaluation purposes~\citep{chen2025pg,qian2025webgrapheval,zhang2025xbound}. 
Table~\ref{tab:related_work_comparison} summarizes the key distinctions between these approaches and GUITAR. 
While these methods offer graph-based perspectives on GUI agent behavior, they fall short in distinct ways: \citet{chen2025pg} rely on perceptual-level similarity for state merging, risking state explosion. \citet{qian2025webgrapheval} target cross-model behavioral comparison rather than failure diagnosis. Their graph nodes lack failure statistics, precluding bottleneck identification.

To our knowledge, GUITAR is the first diagnostic framework to construct a \textbf{globally aggregated STG} across trajectories, where each node represents a shared functional state annotated with diagnostic statistics, directly linking failure localization to inference-time improvement.

\section{Analysis of Step-Level Evaluation}
We analyze why step-level evaluation fails to provide reliable diagnostics in GUI environments. Our analysis reveals two fundamental limitations: (1)  high-frequency states dominate overall SR regardless of agent competence at low-frequency states; and (2)  scalar aggregation collapses transition identity and fails to localize failures to specific states or transitions.

\begin{figure}[b]
    \centering
    \includegraphics[width=0.89\linewidth]{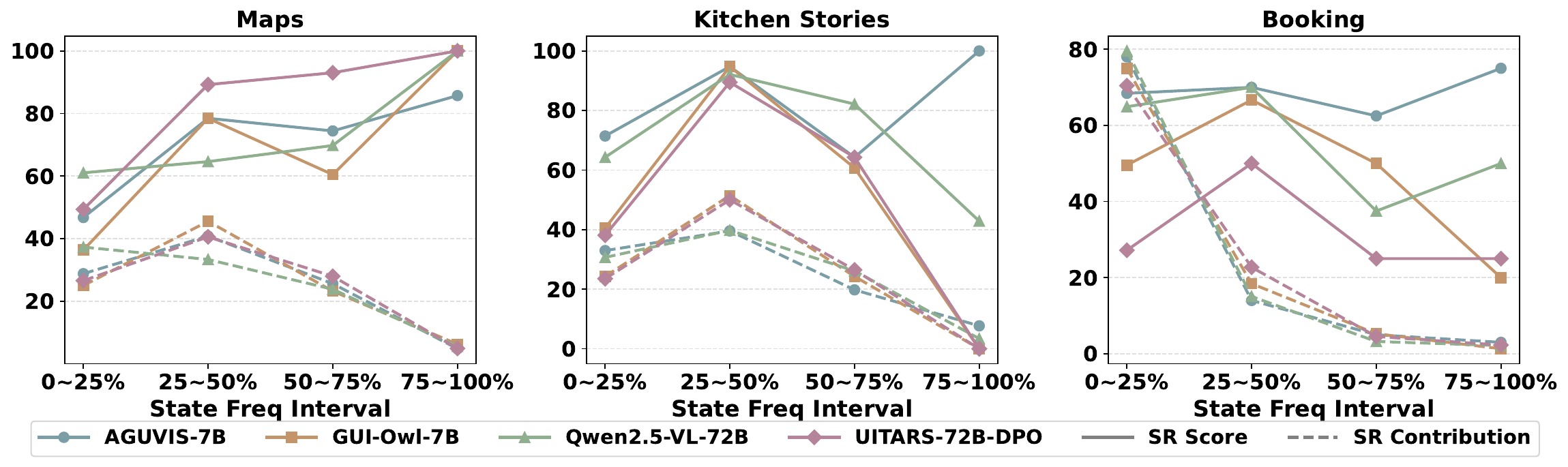}
    \caption{Analysis of performance and SR contribution across state frequency intervals. We observe that long-tail states contribute marginally to the overall SR, regardless of their individual performance.}
    \label{fig:main_result_freq_contribution}
    \vspace{-1ex}
\end{figure}

\subsection{Performance Heterogeneity and Contribution Imbalance}

We analyze the distribution of states and their contributions to overall SR by partitioning states into four quartiles based on visitation frequency. 
We compute the SR within each frequency interval and measure each interval's contribution to the overall SR. 
As illustrated in Figure~\ref{fig:main_result_freq_contribution}, despite non-monotonic performance across frequency intervals, contributions to overall SR are highly skewed: high-frequency states dominate due to their prevalence, whereas low-frequency states contribute marginally regardless of their actual success rates. 

\begin{figure}[t]
  \centering
  \begin{minipage}[b]{0.46\textwidth}
    \captionof{table}{Data ratios across four state-frequency intervals. The 0--25\% interval represents the top 25\% of states ranked by visitation frequency.}
    \label{tab:comprehensive_analysis}
    \centering\small
    \renewcommand{\arraystretch}{1}
    \setlength{\tabcolsep}{1mm}
    \begin{tabular}{l cccc}
        \toprule
        \multirow{2}{*}{\textbf{App}} & \multicolumn{4}{c}{\textbf{ State Frequency Interval (\%)}} \\
        \cmidrule(lr){2-5}
        & \textbf{0--25} & \textbf{25--50} & \textbf{50--75} & \textbf{75--100} \\
        \midrule
        Maps & 40.10 & 33.85 & 22.40 & 3.65 \\
        % CNN & 70.13 & 14.29 & 9.09 & 6.49 \\
        Kitchen Stories & 36.52 & 33.04 & 24.35 & 6.09 \\
        Booking & 78.08 & 13.70 & 5.48 & 2.74 \\
        \midrule
        Average & 51.57 & 26.86 & 17.41 & 4.16 \\
        \bottomrule
    \end{tabular}
  \end{minipage}
  \hfill 
  \begin{minipage}[t]{0.52\textwidth}
    \centering
    \includegraphics[width=\textwidth]{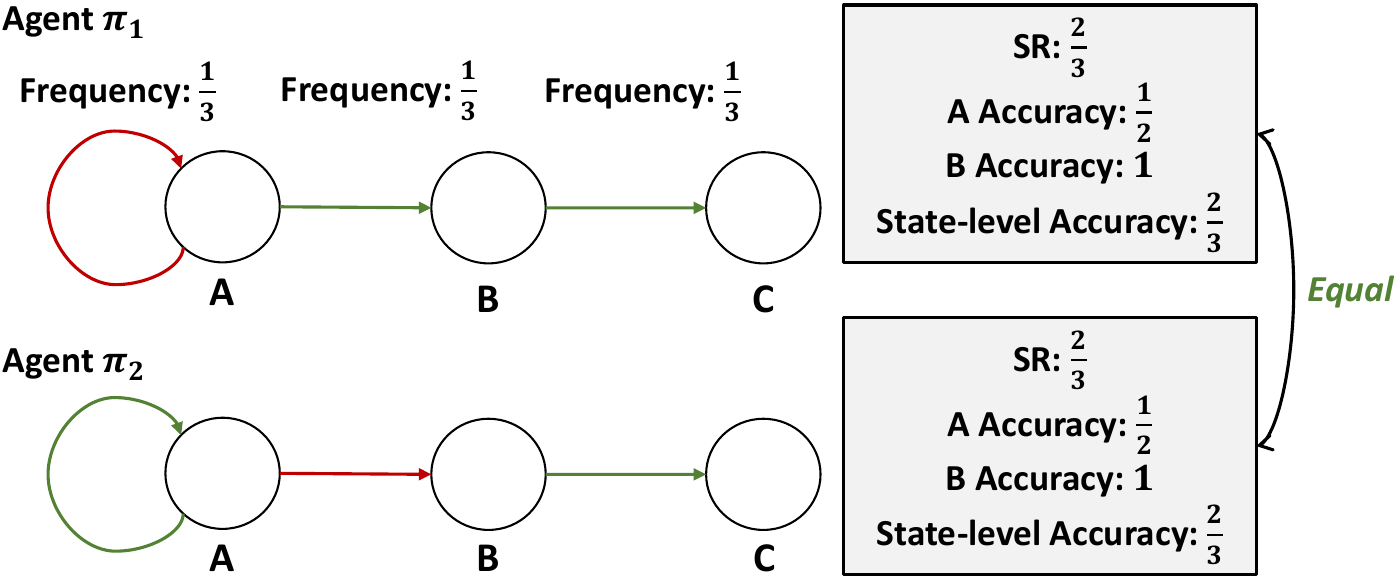}
    \caption{Step-level and state-level aggregation fail to distinguish the failure locations of $\pi_1$ and $\pi_2$. Transition-level analysis reveals distinct bottlenecks.}
    \label{fig:toy_aggregation}
  \end{minipage}
\end{figure}

We rank states by decreasing visitation frequency and divide them into four quartiles, treating the top 25\% as high-frequency states.
As shown in Table~\ref{tab:comprehensive_analysis}, the top 25\% of states account for 51.57\% of all interactions on average, confirming the long-tailed nature of GUI trajectories.

\subsection{Analysis of Scalar Aggregation}\label{sec:non_equivalence}

While the previous analysis motivates the need for state-level abstraction, a natural question remains: Is scalar aggregation sufficient for failure diagnosis?
We show that even with perfect state abstraction, aggregation remains insufficient, as it collapses the structure of interactions and obscures where failures occur. 
Consider an environment with abstract states $\{A,B,C\}$ and transitions
$A \rightarrow A \rightarrow B \rightarrow C$, where $A \rightarrow A$ represents repeated interaction within the same state. 
Define two agents:
\begin{itemize}
\item $\pi_1$ frequently fails at the self-transition $A \rightarrow A$, but reliably executes $A \rightarrow B$ and $B \rightarrow C$.
\item $\pi_2$ reliably executes $A \rightarrow A$, but frequently fails at the transition $A \rightarrow B$.
\end{itemize}

Figure~\ref{fig:toy_aggregation} illustrates that aggregation-based metrics collapsing transition identity assign identical scores to $\pi_1$ and $\pi_2$, despite their failures occurring at different structural locations. 
Therefore, scalar aggregation alone cannot localize errors, which necessitates a novel diagnostic framework. 
\textbf{These results highlight a fundamental limitation of step-level evaluation and motivate finer-grained diagnostic frameworks that can localize such hidden failures.}

% !TeX root = ../neurips_2026.tex

\section{Method}

We introduce GUITAR, a diagnostic framework that analyzes GUI agent behavior through an STG abstraction. 
The STG serves as a reusable diagnostic substrate: constructed once from collected trajectories, it provides a stable structural representation over which multi-level failure analysis is performed. 
Conventional metrics such as SR remain unchanged, whereas GUITAR adds structural diagnostics to localize failures at the state and transition level.

\subsection{State Transition Graph Construction}

GUI environments exhibit high visual variability. Functionally equivalent screens often differ significantly in layout and content. This variability hinders the construction of a consistent graph from raw observations. 
Direct alignment of visually distinct screens across trajectories based on image similarity is challenging. 
It is therefore necessary to bypass visual differences and abstract raw screens into semantically meaningful states.
\begin{figure}
    \centering
    \includegraphics[width=0.8\linewidth]{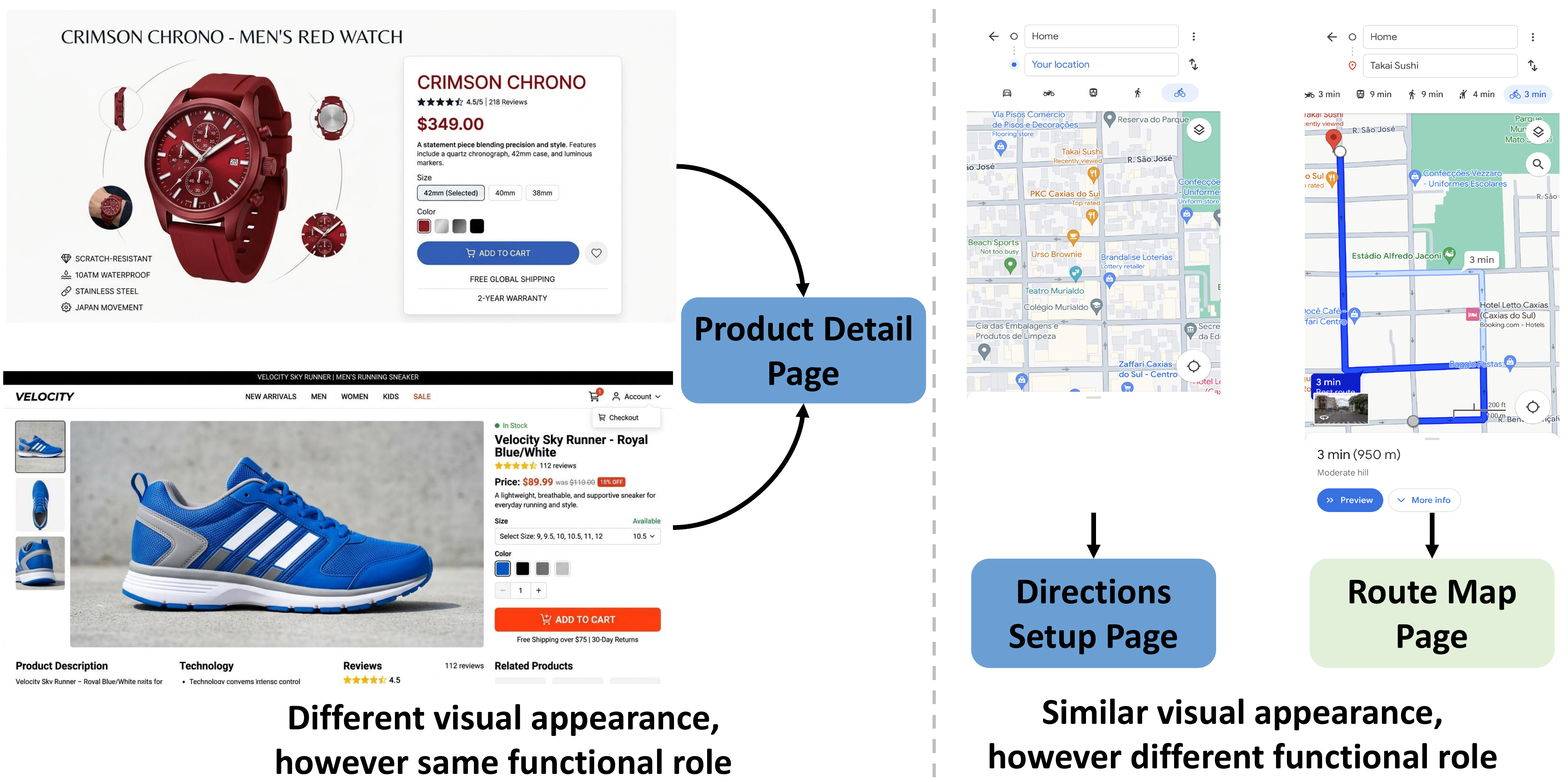}
    \caption{Examples of screen-to-functional-role mappings. 
    Left: visually different screens mapped to the same functional role (item purchasing). 
    Right: visually similar screens mapped to different functional roles, e.g., Directions Setup Page vs. Route Map Page.}
    \label{fig:functional_role_state}
\end{figure}

\paragraph{State Definition.} 
We define a state by what a screen does in the task, not what it looks like. Two screens are considered the same state if they serve the same functional role, enabling the same task-relevant action toward the same goal, regardless of their visual differences. 
As illustrated in Figure~\ref{fig:functional_role_state}, visually different screens can share the same functional role (e.g., both enabling item purchasing), while visually similar screens can represent distinct states (e.g., a Directions Setup Page vs. a Route Map Page).
Let $x$ denote a screen observation, $a$ denote an execution action, and $g$ denote the task goal. 
The mapping $\phi(x, a, g) \rightarrow s$ assigns a raw screen to an abstract state $s$ representing its functional role in the workflow.
Two screen observations $x_i$ and $x_j$ are assigned to the same state if $\phi(x_i, a, g) = \phi(x_j, a, g)$, regardless of their visual differences.

% We define a \emph{\textbf{state}} as a functional-semantic abstraction that captures the role of a screen in the task execution process, rather than its visual appearance.
% Let $x$ denote a screen observation, $a$ denote an execution action, and $g$ denote the task goal. 
% Then $\phi(x, a, g) \rightarrow s,$ which maps a raw screen to an abstract state $s$ representing its functional role in the workflow. 
% The abstraction function $\phi$ operates on two complementary criteria:
% (1) Functional Purpose: what task-relevant action the screen enables or completes under the goal $g$ (e.g., ``selecting a destination'' vs. ``confirming a booking'');
% (2) Behavioral Equivalence: whether the set of task-relevant actions available on $x_i$ and $x_j$ is functionally interchangeable in the context of the trajectory.
% Two $x_i$ and $x_j$ are assigned to the same state if $\phi(x_i, a, g) = \phi(x_j, a, g)$, regardless of their visual differences. 
% The concrete implementation of $\phi$ via VLM-based semantic captioning and state assignment is detailed in Appendix~\ref{sec:stg_construction}.

\paragraph{Graph Construction.}

Given a set of trajectories with step-level correctness annotations in the same app or website, GUITAR constructs an STG to model agent behavior. 
The construction process consists of two stages:

(1) State Abstraction. Each trajectory is first converted into a sequence of abstract states by applying the abstraction function $\phi(x, a, g)$ to every observed screen. 
This produces a state-level trajectory that captures the functional progression of the task.

(2) Graph Aggregation.
We then merge state sequences across trajectories to form a unified graph. 
Each state corresponds to a node, and transitions between consecutive states define directed edges.

The resulting STG captures both 
(i) the structural patterns of interaction and 
(ii) the distribution of successes and failures over states and transitions. 
Importantly, this abstraction enables the alignment of semantically equivalent screens across diverse trajectories. 
Detailed construction procedures are provided in Appendix~\ref{sec:stg_construction}.
Overall, the STG provides a structured representation of agent behavior, serving as the foundation for our multi-level analysis.

\subsection{Multi-level Behavioral Analysis}

Once the STG is constructed, GUITAR enables multi-level behavioral analysis by examining agent performance at different levels of structural granularity. 
These levels form a hierarchy of diagnostic resolution, where each level addresses limitations of the previous one. 
% Table~\ref{tab:multi_level_metrics} summarizes the role of each metric in the diagnostic pipeline.

\paragraph{State-level Analysis.}
At the state level, the STG aggregates execution outcomes over abstract GUI states. 
For each state $s \in S$, we define the state-level success rate as:
\begin{equation}
C(s) = \frac{ \text{successful executions at } s}{ \text{total visits to } s}.
\end{equation}

This aggregation localizes failures to specific interaction stages. 
By examining $C(s)$ across states, GUITAR identifies failure-prone regions in the workflow, i.e., states where execution frequently breaks down. 
Compared to step-level metrics, which mix heterogeneous contexts, state-level analysis provides a structured view of where failures concentrate. 
However, state-level aggregation merges multiple interaction outcomes within a state and therefore cannot distinguish between different types of failures occurring in that state.

\paragraph{Transition-level Analysis.}
To overcome this limitation, GUITAR further analyzes behavior at the transition level. 
Each directed edge $(s_i \rightarrow s_j)$ in the STG represents a concrete interaction step that moves the interface between functional stages. 
We define the transition-level success rate as:
\begin{equation}
T(s_i, s_j) = \frac{\text{successful executions of } (s_i \rightarrow s_j)}{\text{attempts to execute } (s_i \rightarrow s_j)}.
\end{equation}

By measuring $T(s_i, s_j)$ over all transitions, GUITAR pinpoints failure-prone transitions, corresponding to specific decision points where agents consistently make errors. 
Unlike state-level aggregation, transition-level analysis isolates individual decisions and preserves their structural dependencies. 
This enables root cause analysis: failures can be attributed to specific transitions rather than entire states, revealing recurring error patterns tied to navigation or decision processes. 

Together, state-level metrics identify bottleneck regions, while transition-level metrics attribute errors to specific decision points, enabling a hierarchical analysis that transforms evaluation from coarse outcome measurement to a structured diagnosis of agent behavior.

% As summarized in Table~\ref{tab:multi_level_metrics}, state-level metrics identify bottleneck regions, while transition-level metrics attribute errors to precise decision points. 
% This hierarchical analysis transforms evaluation from outcome measurement to structured diagnosis of agent behavior.
% \begin{table}[t]
%     \centering\small
%     \renewcommand{\arraystretch}{1.2}
%     \setlength{\tabcolsep}{1mm}
%     \caption{Multi-level diagnostic roles of evaluation metrics.}
%     \label{tab:multi_level_metrics}
%     \begin{tabular}{ll}
%     \toprule
%     \textbf{Metric}  & \textbf{Diagnostic Role} \\
%     \midrule
%     SR  & Coarse-grained step outcome  \\
%     Task SR  & Coarse-grained task-level outcome \\
%     $C(s)$  & State-level failure localization \\
%     $T(s_i, s_j)$ & Transition-level error attribution \\
%     \bottomrule
%     \end{tabular}
%     \vspace{-2ex}
% \end{table}

% !TeX root = ../neurips_2026.tex
\section{Experiments}
% This section empirically tests three hypotheses derived from the frequency-induced evaluation bias formalized in Section~\ref{sec:frequency_bias}. 

In this section, we validate GUITAR through failure-concentration analysis, agent-specific bottleneck patterns, bottleneck-guided inference, held-out robustness tests, and STG quality evaluation.

\subsection{Datasets and Experimental Setup}\label{sec:dataset}

We evaluate eight GUI agents~\citep{qin2025ui,ye2025mobile,xu2024aguvis,wu2024atlas,bai2025qwen25vltechnicalreport} on 3 apps from AndroidControl~\citep{li2024effects} 
and 3 websites from Mind2Web~\citep{deng2023mind2web}; apps and websites are selected by the number of trajectories in each dataset. 

\begin{figure}[t]
    \centering
    \includegraphics[width=0.95\linewidth]{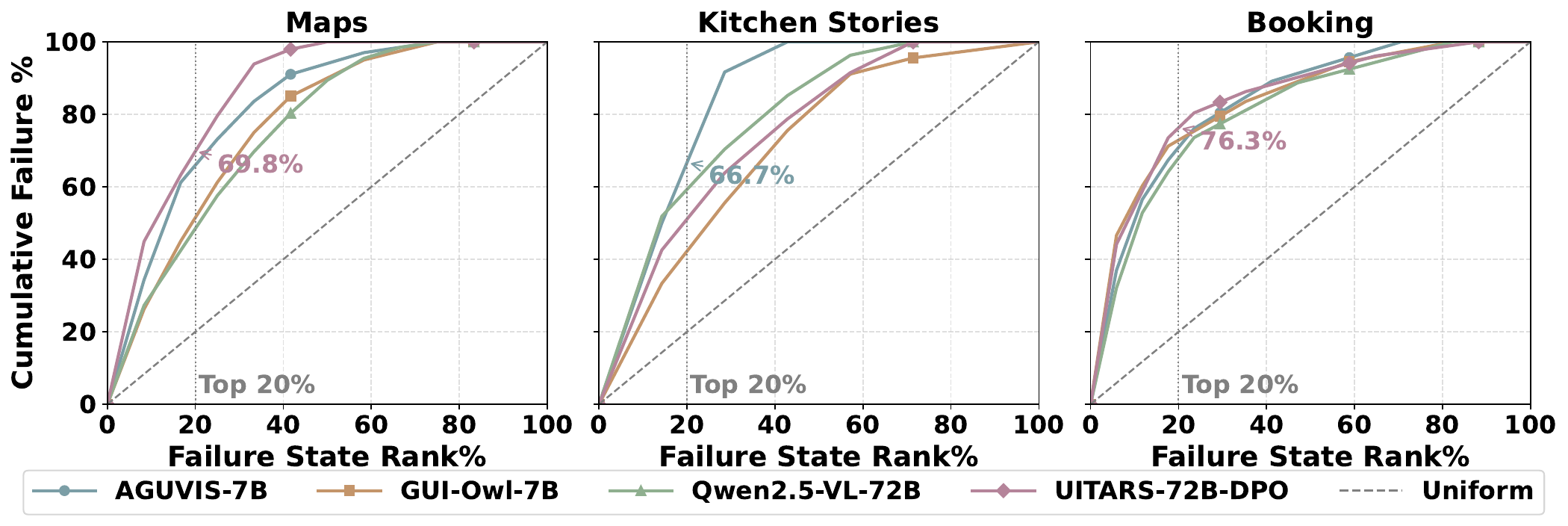}
    \caption{Cumulative distribution of failures across states ranked by failure frequency. A small fraction of states accounts for a disproportionately large portion of total errors, indicating strong failure concentration.}
    \label{fig:main_result_failure_distribution}
    % \vspace{-1ex}
\end{figure}

\subsection{Failure Structure Beyond Aggregate Metrics}\label{sec:failure_structure}

% We leverage the STG to uncover structured failure patterns that are obscured by step-level metrics. Our analysis reveals that agent failures exhibit clear regularities across states and transitions, rather than occurring uniformly.
We leverage the STG to uncover structured failure patterns that are obscured by step-level metrics. Our analysis reveals that agent failures exhibit clear regularities across states and transitions.

\paragraph{Failure Concentration Across States.}
We first examine where failures occur by analyzing their distribution over states. Specifically, we rank states by their failure counts and compute the cumulative proportion of errors. As shown in Figure~\ref{fig:main_result_failure_distribution}, failures are highly concentrated: on average, $60.4\%$ (95\% CI: [36.1\%, 76.2\%]) of failures concentrate in the 20\% of states across all evaluation settings (8 models $\times$ 6 tasks). This concentration significantly deviates from a uniform distribution ($\chi^2$ test, all $p < 0.001$), and a mean Gini coefficient of $0.584$ further confirms that failures are highly localized rather than evenly spread. These results consistently indicate that agent failures are not uniformly distributed but localized to a small set of \textbf{bottleneck states}.

\begin{figure}[t]
    \centering
    \includegraphics[width=0.92\linewidth]{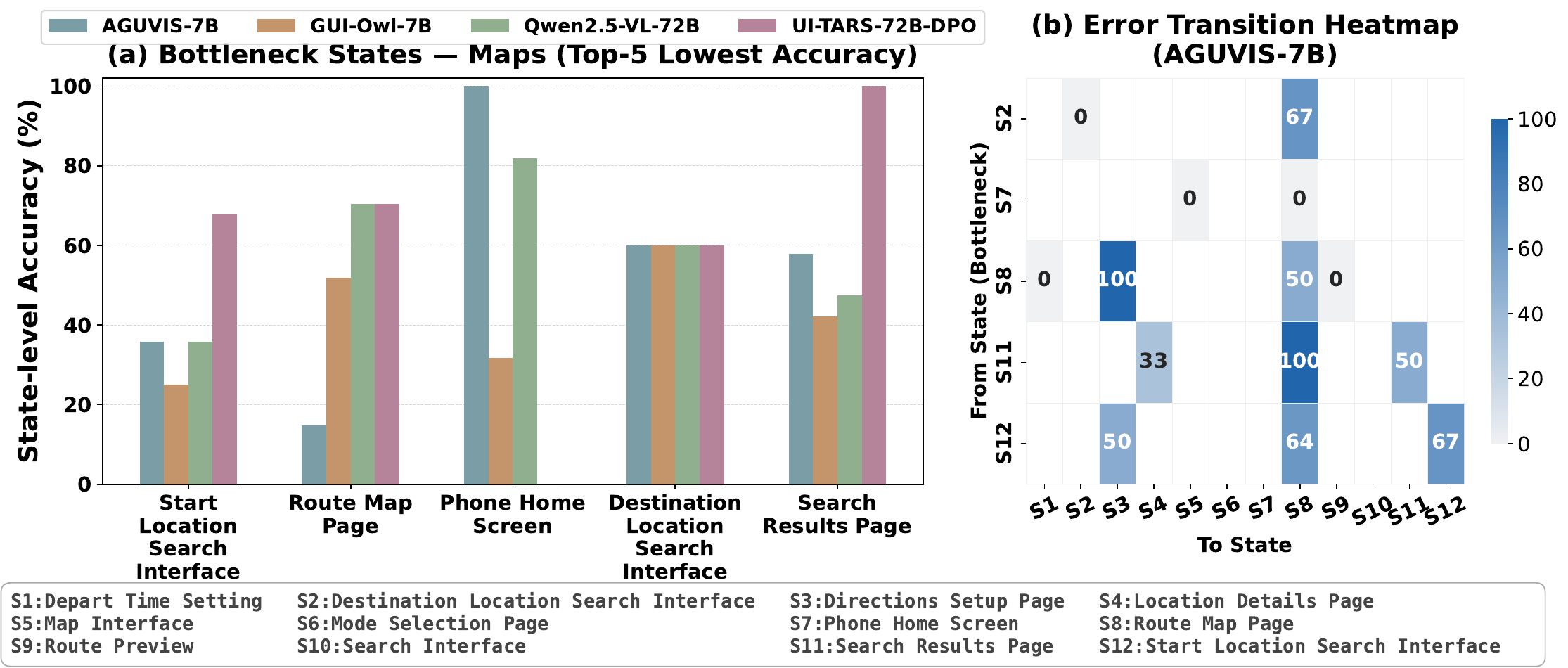}
    % \vspace{-1ex}
    \caption{Analysis of bottleneck states in \textit{Maps}: (a) state-level accuracy of the five most challenging states across all models, and (b) error transition rates originating from bottleneck states (\textit{Aguvis-7B}). GUITAR clearly reveals concentrated failure patterns at both the state and transition levels.}
    \label{fig:maps_bottleneck_combined}
\end{figure}

\paragraph{State-Level Performance Heterogeneity.}
We next investigate how agent performance varies across states. 
Figure~\ref{fig:maps_bottleneck_combined} (a) shows state-wise success rates on the Maps application. 
We observe substantial heterogeneity: (1) individual agents exhibit large performance variation across states, and (2) different agents perform inconsistently on the same state. 
This demonstrates that aggregate step-level metrics obscure important differences in agent capabilities.

\paragraph{Transition-Level Failure Structure.}
While state-level analysis reveals where failures occur, it does not explain how they arise. 
We therefore extend our analysis to transitions between states. 
Figure~\ref{fig:maps_bottleneck_combined} (b) presents a transition-level error heatmap, where each entry corresponds to the failure frequency of a state transition. 
We observe clear structural patterns: a small number of transitions consistently exhibit high error rates. 
This indicates that failures are tied to specific decisions rather than isolated states.
Such transition-level bottlenecks cannot be captured by step-level or state-level analysis alone, highlighting the importance of modeling state transitions for failure diagnosis.

\subsection{Effectiveness of Identified Bottleneck States}\label{sec:effectiveness_of_bottleneck_states}

\begin{table*}[t]
\centering
\caption{SR improvement ($\Delta$SR, pp) of targeted (\textbf{Tgt.}) vs. uniform (\textbf{Uni.}) guidance over the original baseline.
Tgt. denotes GUITAR-guided inference targeting bottleneck states (Threshold = 0.6); 
Uni. denotes uniform guidance applied to all states (Threshold = 0).
Best results per model are \textbf{bolded}.}
\label{tab:targeted_vs_uniform}
\resizebox{\textwidth}{!}{%
\begin{tabular}{l cc cc cc cc cc cc cc}
\toprule
\multirow{2}{*}{\textbf{Model}} 
& \multicolumn{2}{c}{\textbf{Maps}} 
& \multicolumn{2}{c}{\textbf{Kitchen}} 
& \multicolumn{2}{c}{\textbf{CNN}} 
& \multicolumn{2}{c}{\textbf{Booking}} 
& \multicolumn{2}{c}{\textbf{Carmax}} 
& \multicolumn{2}{c}{\textbf{Kayak}} 
& \multicolumn{2}{c}{\textbf{Avg.}} \\
\cmidrule(lr){2-3}\cmidrule(lr){4-5}\cmidrule(lr){6-7}
\cmidrule(lr){8-9}\cmidrule(lr){10-11}\cmidrule(lr){12-13}\cmidrule(lr){14-15}
& Uni. & Tgt. 
& Uni. & Tgt. 
& Uni. & Tgt. 
& Uni. & Tgt. 
& Uni. & Tgt. 
& Uni. & Tgt. 
& Uni. & Tgt. \\
\midrule
Qwen2.5-VL-7B   
& -3.5  & \textbf{+1.0}  
& +8.3  & \textbf{+4.6}  
& -7.6  & \textbf{+2.6}  
& -12.3 & \textbf{+1.4}  
& -13.1 & \textbf{-0.7}  
& -17.3 & \textbf{+1.1}  
& -7.6  & \textbf{+1.7}  \\

Qwen2.5-VL-72B  
& 0.0   & \textbf{+0.6}  
& +3.0  & \textbf{+1.8}  
& -17.9   & 0.0   
& -9.6  & \textbf{+1.4}  
& -9.2  & \textbf{+3.9}  
& +3.9  & +3.9  
& -5.0  & \textbf{+2.0}  \\

OS-Atlas-7B     
& 0.0   & \textbf{+3.5}  
& -1.4  & \textbf{+2.3}  
& +1.3  & \textbf{+2.5}  
& +0.4  & \textbf{+0.3}  
& +3.0  & \textbf{+8.0}  
& +9.5  & +9.5  
& +2.1  & \textbf{+4.4}  \\

GUI-Owl-7B      
& +5.5  & \textbf{+6.0}  
& \textbf{+7.5}  & +1.5  
& -7.6  & \textbf{+1.3}  
& +1.0  & \textbf{+3.1}  
& \textbf{-0.9}  & -8.7  
& -2.7  & \textbf{-6.0}  
& \textbf{+0.5}  & -0.5  \\

GUI-Owl-32B     
& +8.5  & +8.5  
& +11.2 & \textbf{+12.5} 
& -6.4  & \textbf{+2.5}  
& +0.5  & \textbf{+5.3}  
& -4.8  & \textbf{+6.3}  
& -11.5 & \textbf{-12.0} 
& -0.4  & \textbf{+3.9}  \\

Aguvis-7B       
& +1.5  & \textbf{+5.0}  
& 0.0   & 0.0   
& -3.7  & \textbf{+1.3}  
& -4.1  & \textbf{+0.7}  
& -12.4 & \textbf{0.0}   
& -19.4 & \textbf{0.0}   
& -6.4  & \textbf{+1.2}  \\

UI-TARS-1.5-7B   
& -16.1 & \textbf{+18.7} 
& -6.5  & \textbf{+21.2} 
& +8.8  & \textbf{+6.3}  
& -3.4  & \textbf{+2.1}  
& -1.2  & \textbf{0.0}   
& \textbf{+1.6}  & -2.8  
& -2.8  & \textbf{+7.6}  \\

UI-TARS-72B-DPO  
& -10.2 & \textbf{+4.3}  
& -3.0  & \textbf{+12.6} 
& -9.9  & \textbf{-6.1}  
& -4.8  & \textbf{-1.3}  
& -7.2  & \textbf{+0.7}  
& +0.5  & \textbf{0.0}   
& -5.8  & \textbf{+1.7}  \\

\midrule
\textbf{Avg.}   
& -1.8  & \textbf{+5.9}  
& +2.4  & \textbf{+7.1}  
& -5.4  & \textbf{+1.3}  
& -4.0  & \textbf{+1.6}  
& -5.7  & \textbf{+1.2}  
& -4.4  & \textbf{-0.5}  
& -3.2  & \textbf{+2.8}  \\
\bottomrule
\end{tabular}%
}
\end{table*}

To validate that the bottleneck states identified by GUITAR reflect genuine capability deficiencies, we inject targeted guidance at identified bottleneck states during inference and compare against uniform guidance across all states as a control.  
Table~\ref{tab:targeted_vs_uniform} shows that targeted guidance improves SR by 2.8 percentage points on average, while uniform guidance reduces SR by 3.2 percentage points, confirming that the improvement stems from precise bottleneck identification rather than the mere introduction of additional information.

Table~\ref{tab:bottleneck_mitigation} reveals the underlying mechanism: bottleneck state SR improves by 18.1 percentage points on average under targeted guidance, directly demonstrating that the identified bottlenecks are successfully mitigated.
Critically, a strong task-level correspondence exists between bottleneck mitigation and SR improvement: tasks with higher $\Delta C(s)$ (Kitchen: +37.5 pp, Maps: +32.6 pp) consistently yield larger SR gains (+7.1 pp, +5.9 pp), while tasks with limited mitigation (Carmax: +5.1 pp, Kayak: +5.6 pp) show little improvement or a decline in SR (+1.2 pp, -0.5 pp). This positive correlation confirms that overall performance gains are driven by bottleneck mitigation.

% However, two failure modes limit improvements in certain cases. First, when the assistant model fails to generate effective sub-goals, bottleneck states remain unmitigated, resulting in negligible SR gains. Second, even when bottlenecks are successfully mitigated, models with weaker instruction-following capability fail to translate bottleneck-level improvements into task-level success, suggesting that more fine-grained guidance mechanisms are needed.

However, three failure modes limit improvements in certain cases. First, during retrieval, not only are bottleneck states retrieved, but some well-performing states are also retrieved, thereby triggering auxiliary reasoning unnecessarily. 
Second, when the assistant model fails to generate valid sub-goals, failures at the bottleneck states cannot be mitigated, resulting in negligible SR gains.
Finally, even when correct sub-goals are generated, in some scenarios the agent fails to take the correct actions, indicating the need for more fine-grained guidance mechanisms.

\begin{table*}[t]
\centering
\caption{Success rates at bottleneck states (\%) before and after targeted guidance across six tasks. Orig.\ and Tgt.\ report the original and guided rates, respectively; the final row reports $\Delta C(s)$ in percentage points.}
\label{tab:bottleneck_mitigation}
\resizebox{\textwidth}{!}{%
\begin{tabular}{lcccccccccccccc}
\toprule
\multirow{2}{*}{\textbf{Model}} 
& \multicolumn{2}{c}{\textbf{Maps}} 
& \multicolumn{2}{c}{\textbf{Kitchen}} 
& \multicolumn{2}{c}{\textbf{CNN}} 
& \multicolumn{2}{c}{\textbf{Booking}} 
& \multicolumn{2}{c}{\textbf{Carmax}} 
& \multicolumn{2}{c}{\textbf{Kayak}} 
& \multicolumn{2}{c}{\textbf{Avg.}} \\
\cmidrule(lr){2-3} \cmidrule(lr){4-5} \cmidrule(lr){6-7} 
\cmidrule(lr){8-9} \cmidrule(lr){10-11} \cmidrule(lr){12-13}
\cmidrule(lr){14-15}
& Orig. & Tgt. 
& Orig. & Tgt. 
& Orig. & Tgt. 
& Orig. & Tgt. 
& Orig. & Tgt. 
& Orig. & Tgt. 
& Orig. & Tgt. \\
\midrule
Qwen2.5-VL-7B   
& 23.6 & 38.3 
& 10.5 & 52.4 
& 0.0  & 33.3 
& 10.0 & 40.0 
& 25.0 & 8.3  
& 14.3 & 14.3 
& 13.9 & 31.1 \\

Qwen2.5-VL-72B  
& 27.8 & 39.3 
& 22.0 & 47.6 
& 0.0  & 0.0  
& 23.5 & 47.1 
& 16.0 & 40.0 
& 14.3 & 14.3 
& 17.3 & 31.4 \\

OS-Atlas-7B     
& 27.9 & 31.2 
& 22.9 & 36.4 
& 27.8 & 44.4 
& 34.2 & 35.5 
& 16.2 & 27.6 
& 28.4 & 49.1 
& 26.2 & 37.4 \\

GUI-Owl-7B      
& 24.1 & 54.0 
& 16.7 & 29.3 
& 30.4 & 36.4 
& 30.7 & 35.6 
& 34.1 & 41.7 
& 0.0  & 0.0  
& 22.7 & 32.8 \\

GUI-Owl-32B     
& 25.4 & 62.8 
& 17.4 & 60.7 
& 26.7 & 34.9 
& 30.2 & 36.1 
& 34.4 & 47.8 
& 16.7 & 16.7 
& 25.1 & 43.2 \\

Aguvis-7B       
& 21.9 & 44.6 
& 6.5  & 50.0 
& 0.0  & 33.3 
& 25.0 & 37.5 
& 0.0  & 0.0  
& 16.7 & 33.3 
& 11.7 & 33.1 \\

UI-TARS-1.5-7B   
& 2.8  & 64.3 
& 6.9  & 65.9 
& 13.6 & 42.1 
& 19.4 & 24.2 
& 17.6 & 16.8 
& 21.4 & 23.7 
& 13.6 & 39.5 \\

UI-TARS-72B-DPO  
& 6.7  & 86.4 
& 5.2  & 65.9 
& 22.2 & 34.9 
& 15.1 & 17.9 
& 15.4 & 16.9 
& 16.2 & 21.7 
& 13.5 & 40.6 \\

\midrule
\textbf{Avg.}   
& 20.0 & 52.6 
& 13.5 & 51.0 
& 15.1 & 32.4 
& 23.5 & 34.2 
& 19.8 & 24.9 
& 16.0 & 21.6 
& 18.0 & 36.1 \\

\midrule
$\Delta C(s)$ (pp)
& \multicolumn{2}{c}{\textbf{+32.6}} 
& \multicolumn{2}{c}{\textbf{+37.5}} 
& \multicolumn{2}{c}{\textbf{+17.3}} 
& \multicolumn{2}{c}{\textbf{+10.7}} 
& \multicolumn{2}{c}{\textbf{+5.1}}  
& \multicolumn{2}{c}{\textbf{+5.6}}  
& \multicolumn{2}{c}{\textbf{+18.1}} \\
\bottomrule
\end{tabular}%
}
\vspace{-2ex}
\end{table*}

\subsection{Robustness and Held-out Validation}
\label{sec:robustness_heldout}
\vspace{-1ex}
\begin{table*}[t]
\centering
\small
\caption{Robustness to manual verification and trajectory reuse on the six primary tasks. Panel A compares fully automatic and manually verified STGs; Panel B reports three-fold held-out guidance using fully automatic STGs.}
\label{tab:robustness_heldout}
\renewcommand{\arraystretch}{1.12}
\setlength{\tabcolsep}{3pt}
\begin{tabular}{l*{14}{c}}
\toprule
\multicolumn{15}{l}{\textbf{Panel A: Fully automatic vs. manually verified STGs}} \\
\midrule
\textbf{Dataset}
& \multicolumn{4}{c}{\textbf{Top-20\% Jaccard}}
& \multicolumn{4}{c}{\textbf{Spearman $\rho$}}
& \multicolumn{3}{c}{\textbf{$|\Delta\mathrm{Gini}|$}}
& \multicolumn{3}{c}{\textbf{$|\Delta\mathrm{Coverage}|$ (pp)}} \\
\cmidrule(lr){2-5}
\cmidrule(lr){6-9}
\cmidrule(lr){10-12}
\cmidrule(lr){13-15}
AndroidControl
& \multicolumn{4}{c}{$0.932 \pm 0.214$}
& \multicolumn{4}{c}{$0.989 \pm 0.029$}
& \multicolumn{3}{c}{$0.017 \pm 0.024$}
& \multicolumn{3}{c}{$1.89 \pm 2.70$} \\
Mind2Web$^{\dagger}$
& \multicolumn{4}{c}{$0.500 \pm 0.136$}
& \multicolumn{4}{c}{$0.942 \pm 0.035$}
& \multicolumn{3}{c}{$0.043 \pm 0.025$}
& \multicolumn{3}{c}{$6.26 \pm 3.00$} \\
\midrule
\multicolumn{15}{l}{\textbf{Panel B: Three-fold trajectory-held-out guidance with fully automatic STGs}} \\
\midrule
\textbf{Metric}
& \multicolumn{2}{c}{\textbf{Maps}}
& \multicolumn{2}{c}{\textbf{Kitchen}}
& \multicolumn{2}{c}{\textbf{CNN}}
& \multicolumn{2}{c}{\textbf{Booking}}
& \multicolumn{2}{c}{\textbf{Carmax}}
& \multicolumn{2}{c}{\textbf{Kayak}}
& \multicolumn{2}{c}{\textbf{Overall}} \\
\cmidrule(lr){2-3}
\cmidrule(lr){4-5}
\cmidrule(lr){6-7}
\cmidrule(lr){8-9}
\cmidrule(lr){10-11}
\cmidrule(lr){12-13}
\cmidrule(lr){14-15}
Mean $\Delta$SR (pp)
& \multicolumn{2}{c}{$+2.98$}
& \multicolumn{2}{c}{$+3.29$}
& \multicolumn{2}{c}{$+0.87$}
& \multicolumn{2}{c}{$+1.20$}
& \multicolumn{2}{c}{$+1.09$}
& \multicolumn{2}{c}{$+1.27$}
& \multicolumn{2}{c}{\textbf{$+1.88$}} \\
Improved / Tied / Decreased
& \multicolumn{2}{c}{$4/3/0$}
& \multicolumn{2}{c}{$5/2/0$}
& \multicolumn{2}{c}{$4/3/0$}
& \multicolumn{2}{c}{$1/4/0$}
& \multicolumn{2}{c}{$2/2/1$}
& \multicolumn{2}{c}{$1/2/2$}
& \multicolumn{2}{c}{\textbf{$17/16/3$}} \\
\bottomrule
\end{tabular}%

\begin{minipage}{\textwidth}
\footnotesize
$^{\dagger}$Jaccard uses exact node-name matching. $|\Delta\mathrm{Coverage}|$ is the absolute change in the failure mass captured by the Top-20\% states. Overall $\Delta$SR averages 36 task--agent pairs drawn from seven agents.
\end{minipage}
\end{table*}

We examine whether GUITAR's conclusions depend on (i) post-hoc manual verification or (ii) trajectory reuse between bottleneck estimation and guided inference.

\paragraph{Automatic vs. Verified STGs.}
We recompute the failure statistics using the fully automatic Stage~5 assignments. As shown in Table~\ref{tab:robustness_heldout} (Panel A), automatic and verified STGs produce strongly correlated failure rankings ($\rho=0.989/0.942$ on AndroidControl/Mind2Web), with small changes in Gini coefficient and Top-20\% failure coverage. The lower exact-name Jaccard on Mind2Web is primarily due to differently named but semantically aligned states. Thus, manual verification improves assignment fidelity but is not necessary to recover the principal bottleneck structure.

\paragraph{Trajectory-Held-out Guidance.}
Beyond the baseline task instruction, STG guidance contains only cross-trajectory functional roles and abstract transitions; it does not reveal the evaluated episode's future trajectory, outcome, or prescribed next action. Nevertheless, because the original result showing a gain of 2.8 percentage points is obtained using overlapping trajectories for STG construction, error estimation, and evaluation, we conduct three-fold evaluation using two thirds of the trajectories for construction and the untouched third for testing. Across the same six tasks, guidance improves SR by 1.88 percentage points over 36 task--agent pairs (17 improved, 16 tied, and 3 decreased; Table~\ref{tab:robustness_heldout}, Panel B). This demonstrates within-task generalization to unseen trajectories; results on three additional tasks are in Appendix~\ref{sec:extended_heldout}.

\subsection{Validation of State Abstraction}\label{sec:validation_of_state_abstraction}

The effectiveness of GUITAR depends on the quality of the constructed state abstractions. 
We evaluate both the semantic coherence of states and the validity of alternative abstraction strategies.

\paragraph{Sensitivity to Abstraction Errors.}
We test whether imperfect state abstraction changes bottlenecks by randomly introducing over-merge and over-split errors into 10\% of state labels. On AndroidControl and Mind2Web, the perturbed graphs retain Top-20\% bottleneck Jaccard overlaps of $0.892$ and $0.855$, failure-rank correlations of $0.961$ and $0.973$, and Top-1 bottleneck survival rates of $0.920$ and $0.962$, respectively. The diagnostic conclusions remain stable under plausible assignment noise.

\paragraph{Reliable and Consistent State Abstraction.}

\begin{wraptable}{r}{0.321\textwidth}
\vspace{-2ex}
\centering\small
\renewcommand{\arraystretch}{0.9}
\setlength{\tabcolsep}{1.4mm}
\caption{Inter-annotator agreement for state abstraction evaluation (5 annotators, 71 states).}
\label{tab:annotation_agreement}
\begin{tabular}{lc}
\toprule
\textbf{Metric} & \textbf{Score} \\
\midrule
Gwet's AC1              & 0.848  \\
Raw Agreement           & 83.1\% \\
\bottomrule
\end{tabular}
\end{wraptable}

To verify that GUITAR's state descriptions are semantically coherent and consistently identifiable, we conduct a human annotation study over 71 states spanning all 6 tasks. 
% For each state, five annotators with computer science backgrounds and prior GUI task experience independently perform binary judgments on five randomly sampled screenshots, indicating whether each screenshot matches the corresponding state description. 
As shown in Table~\ref{tab:annotation_agreement}, raw agreement reaches 83.1\%, and Gwet's AC1 score is 0.848. 
These results confirm that GUITAR's state abstractions are well-defined and reliably recognized by independent annotators across all evaluated tasks, providing a strong foundation for the failure diagnosis analyses that follow. Further details are provided in Appendix~\ref{sec:stg_evaluation}.
% We first assess whether states are semantically coherent and consistently identifiable. 
% For each state, we sample five screenshots and ask five annotators to judge whether each sample matches the corresponding state description. 
% As shown in Table~\ref{tab:annotation_agreement}, agreement is high across all metrics. 
% Raw agreement reaches 83.1\%, and Gwet's AC1 score is 0.8481, indicating \textit{almost perfect} agreement. 
% These results confirm that the states are well-defined and reliably recognized by independent annotators.

% \begin{table}[t]
% \centering\small
% \renewcommand{\arraystretch}{1.2}
% \setlength{\tabcolsep}{1mm}
% \caption{Inter-annotator agreement for state abstraction evaluation 
%          (5 annotators, 71 states).}
% \label{tab:annotation_agreement}
% \begin{tabular}{lcc}
% \toprule
% \textbf{Metric} & \textbf{Score} & \textbf{Interpretation} \\
% \midrule
% Gwet's AC1    & 0.9154 & Almost Perfect \\
% Raw Agreement           & 83.1\% & 95\% CI [72.7\%, 90.1\%] \\
% \midrule
% \multicolumn{3}{l}{\textit{Threshold sensitivity (Gwet's AC1):}} \\
% 60\% threshold        & 0.9154 & Almost Perfect \\
% 70\% threshold        & 0.8521 & Almost Perfect \\
% 80\% threshold        & 0.8481 & Almost Perfect \\
% \bottomrule
% \end{tabular}
% % \vspace{-2ex}
% \end{table}

\begin{table}[t]
\centering\small
\renewcommand{\arraystretch}{0.9}
\setlength{\tabcolsep}{1.8mm}
\caption{Comparison of STGs constructed by different methods, including PageAgent~\citep{chen2025pg}. 
GUITAR achieves a more compact state abstraction while preserving both visual similarity and behavioral consistency.
Ratio denotes the proportion of image pairs with similarity above 70\%. 
Intra and Inter represent the textual similarity within and across transitions, respectively. 
\myuparrow indicates higher is better, and \mydownarrow indicates lower is better. 
\textbf{Bold} denotes the best result.}
\label{tab:transition_consistency}
\begin{tabular}{l l c ccc ccccc}
\toprule

% ── 一级表头 ──────────────────────────────────────────────────
\multirow{2}{*}{\textbf{App}}
& \multirow{2}{*}{\textbf{Method}}
& \multirow{2}{*}{\textbf{States}}
& \multicolumn{3}{c}{\textbf{State Image Similarity}}
& \multicolumn{4}{c}{\textbf{Action Similarity}} \\

\cmidrule(lr){4-6} \cmidrule(lr){7-10}

% ── 二级表头 ──────────────────────────────────────────────────
& 
&
& Mean\myuparrow
& Std\mydownarrow
& Ratio\myuparrow
& Intra\myuparrow
& Inter\mydownarrow
& Gap\myuparrow
& $p$-value \\

\midrule

% ══════════════════════════════════════════════════════════════
% App 1: Kitchen Stories
% ══════════════════════════════════════════════════════════════
\multirow{3}{*}{Kitchen Stories}
& ImageHash & 12
    & 75.4          & 14.8          & 57.1
    & 54.0          & 37.4          & 16.6          & \textbf{4.7e-10} \\

& PageAgent & 45
    & \textbf{82.2}          & \textbf{9.7}           & 83.3
    & 78.6          & \textbf{29.0}          & 27.3          & 6.1e-06 \\

& \textbf{GUITAR} & 7
    & 80.9 & \textbf{9.7}  & \textbf{85.7}
    & \textbf{75.8} & 41.0 & \textbf{34.8} & 5.2e-09 \\

\midrule

% ══════════════════════════════════════════════════════════════
% App 2: Maps
% ══════════════════════════════════════════════════════════════
\multirow{3}{*}{Maps}
& ImageHash & 12
    & 68.2          & 12.1          & 37.9
    & 42.1          & 42.7          & $-$0.7        & 0.6679 \\

& PageAgent & 63
    & \textbf{71.5} & 13.4  & \textbf{48.0}
    & 37.0          & 39.9 & $-$2.9           & 0.8907 \\

& \textbf{GUITAR} & 12
    & 70.6          & \textbf{9.5}           & 41.2
    & \textbf{60.3} & \textbf{38.4}          & \textbf{21.9} & \textbf{0.0000} \\

\midrule

\multirow{3}{*}{\makecell[l]{CNN}}
& ImageHash & 9
    & 68.1          & 15.4          & 35.7
    & 60.1          & 35.4          & 24.7          & \textbf{3.86e-9} \\

& PageAgent & 38
    & 69.4          & \textbf{12.6}           & 30.8
    & 45.1          & 33.0          & 12.1          & 0.1727 \\

& \textbf{GUITAR} & 7
    & \textbf{71.9} & 15.1  & \textbf{53.9}
    & \textbf{65.7} & \textbf{28.5} & \textbf{37.2} & 0.0000 \\

\midrule

\multirow{3}{*}{Booking}
& ImageHash & 19
    & 80.4         & 9.2          & 88.5
    & 43.4          & 42.4          & 1.0          & 0.3732 \\

& PageAgent & 12
    & \textbf{82.6}          & \textbf{8.1}          & \textbf{100.0}
    & 29.0          & \textbf{22.9}          & 6.1          & 0.0907 \\

& \textbf{GUITAR} & 17
    & 77.8          & 8.9          & 82.4
    & \textbf{65.6} & 35.2 & \textbf{30.5} & \textbf{0.0000} \\
\midrule

\multirow{3}{*}{Carmax}
& ImageHash & 22
    & \textbf{88.9}          & 8.7          & \textbf{93.8}
    & 45.2          & 33.8          & 11.4        & 1.30e-8 \\

& PageAgent & 12
    & 84.6 & \textbf{8.6}  & 90.9
    & 39.1          & \textbf{28.9} & \textbf{10.2}           & 0.0302 \\

& \textbf{GUITAR} & 12
    & 82.3          & 9.3           & 89.0
    & \textbf{47.0} & 36.8 & \textbf{10.2} & \textbf{2.95e-10} \\

\midrule

\multirow{3}{*}{Kayak}
& ImageHash & 8
    & 84.2         & 8.1          & 92.5
    & 38.0          & 32.5          & 5.5          & 1.09e-4 \\

& PageAgent & 7
    & \textbf{87.4}          & 9.3          & \textbf{100.0}
    & 29.7          & \textbf{18.4}          & \textbf{11.3}          & 0.1289 \\

& \textbf{GUITAR} & 13
    & 78.9          & \textbf{5.7}          & 92.3
    & \textbf{42.9} & 35.3 & 7.6 & \textbf{1e-12} \\

% \midrule

% ══════════════════════════════════════════════════════════════
% 平均行
% ══════════════════════════════════════════════════════════════
% \multirow{3}{*}{\textbf{Average}}
% & ImageHash & 12.67
%     & 71.38          & 12.04          & 61.17
%     & 46.48          & 40.84          & 5.63          & -- \\

% & PageAgent & 40
%     & \textbf{78.75}          & 10.41           & \textbf{77.11}
%     & 48.20          & \textbf{30.57}          & 17.63          & -- \\

% & GUITAR & 12
%     & 74.67 & \textbf{9.34} & 69.74
%     & \textbf{67.64} & 38.23 & \textbf{29.48} & -- \\

\bottomrule
\vspace{-4ex}
\end{tabular}%

\end{table}

\paragraph{Limitations of Similarity-Based State Abstraction}

A natural alternative is to construct states via visual or textual similarity clustering. 
We consider two representative baselines: (i) low-level similarity clustering using image hashing and sentence embeddings, and (ii) a VLM-based merging strategy~\citep{chen2025pg}\footnote{The Mind2Web dataset contains trajectories with missing screenshots. When PageAgent encounters a missing image during the merging process, it terminates the merging of that trajectory early, resulting in fewer nodes being merged and thus an artificially reduced node count. This partially explains the lower node counts observed for PageAgent on Mind2Web tasks.}. 
As shown in Table~\ref{tab:transition_consistency}, the three methods exhibit distinct trade-offs. Note that GUITAR constructs the STG used in this comparison without manual correction, ensuring a fair evaluation of the automated pipeline's intrinsic abstraction capability.

ImageHash produces a compact graph (14 states) but achieves the lowest action consistency (gap: 9.75), indicating that low-level features are insufficient for grouping functionally related screens.

PageAgent generates a substantially larger number of states (30), achieving the highest image similarity (Mean: 79.6) but limited behavioral consistency (gap: 10.7), suggesting that VLM-based similarity merging captures visual resemblance but fails to preserve functional structure and results in over-fragmented state representations.

In contrast, GUITAR produces a similarly compact graph (11 states) while achieving the highest intra-transition action similarity (60.1) and the largest intra-inter gap (24.2), demonstrating that functionally grounded abstraction better separates behaviorally distinct states.

These results highlight that similarity-based methods capture superficial resemblance but fail to preserve behavioral consistency, underscoring the necessity of functionally grounded state abstractions for reliable failure diagnosis. Detailed evaluation implementations are provided in Appendix~\ref{sec:abstraction_evaluation}.

\paragraph{Cost Analysis.} 

\begin{wraptable}{r}{0.4\textwidth}
\vspace{-2.5ex}
\centering\small
\renewcommand{\arraystretch}{0.9}
\setlength{\tabcolsep}{1mm}
\caption{VLM call efficiency and graph compactness comparison on AndroidControl.}
\label{tab:vlm_calls}
\begin{tabular}{lcc}
\toprule
\textbf{Method} & \textbf{VLM Calls} & \textbf{Avg.\ States ($N$)} \\
\midrule
PageAgent & 1{,}167 & $\approx$49 \\
GUITAR    & 933     & $\approx$9 \\
\midrule
Reduction & 20.1\%  & 82.6\%      \\
\bottomrule
\end{tabular}
% \vspace{-2ex}
\end{wraptable}

We compare the computational cost of GUITAR and PageAgent by the number of VLM calls, which dominates overall overhead. 
% Let $T$ denote the number of observations and $N_t$ the number of existing states at step $t$. 
% PageAgent compares each new observation against all existing states at every step, yielding $\text{calls} = O\left(\sum_{t=1}^{T} N_t\right).$ calls, reducing to $O(NT)$ in practice when states stabilize. 
% In contrast, GUITAR decouples abstraction from graph construction: the abstraction stage requires one VLM call per observation $O(T)$, and the global merging stage requires a single call regardless of $N$, giving an overall complexity of $O(T)$. 
% On AndroidControl, this translates to 933 VLM calls for GUITAR versus 1167 for PageAgent, a 20.1\% reduction, while also producing a more compact state representation ($N\approx9$ vs. $N\approx49$), confirming that decoupling abstraction from graph construction eliminates redundant pairwise comparisons and achieves better scalability.
As shown in Table~\ref{tab:vlm_calls}, on AndroidControl, GUITAR requires 933 VLM calls compared to 1,167 for PageAgent, achieving a 20.1\% reduction in computational cost. Concurrently, GUITAR produces a substantially more compact state representation ($N\approx9$ vs. $N\approx49$, an 82.6\% reduction in graph size). 
These two improvements are not independent: by leveraging functional page semantics to decouple state abstraction from graph construction, GUITAR avoids the redundant pairwise comparisons inherent in similarity-based merging approaches, where the number of required comparisons grows quadratically with the number of trajectories. 
\section{Conclusion}

% Aggregate metrics can create a competence illusion by masking failure-prone states and transitions in GUI interaction workflows. 
% To address this limitation, we propose GUITAR, which models execution trajectories as a STG and enables multi-level diagnosis of GUI agent failures. 
% By using semantic guidance to map screenshots to unique states, GUITAR distinguishes between high- and low-frequency scenarios. 
% Our empirical results show GUITAR exposes concentrated failure patterns invisible to conventional metrics, proving the necessity of state- and transition-aware diagnostics for faithful GUI agent assessment.

% We present GUITAR, a state-centric framework that diagnoses GUI agent failures through structured analysis of STGs. 
% By mapping screens to functional states based on behavioral equivalence, GUITAR enables systematic failure analysis across both state and transition levels. 
% GUITAR reveals that failures concentrate on a small set of bottleneck states. Leveraging these bottleneck states for inference-time guidance yields consistent performance improvements of +2.8\%.
% These results underscore that modeling the structure of evaluation outcomes is essential for understanding GUI agent reliability and driving principled improvement.

We present GUITAR, a state-centric framework that reframes GUI agent evaluation as a structured analysis problem over STGs. 
By abstracting screens into functional states based on behavioral equivalence, GUITAR enables systematic diagnosis of agent failures at both the state and transition levels, going beyond step-level metrics.
GUITAR reveals that failures concentrate on a small set of bottleneck states. 
Bottleneck-targeted guidance improves SR by 2.8 percentage points and retains an average gain of 1.88 percentage points across 7 agents under three-fold trajectory-held-out evaluation with fully automatic STGs.
These findings establish structure-aware evaluation as an effective framework for identifying execution bottlenecks and guiding targeted improvements in mobile and web GUI agents.

% !TeX root = ../neurips_2026.tex

% \textbf{(1) Trajectory scale and STG coverage.} GUITAR constructs STGs from ground-truth trajectories, and the quality of the resulting graph is inherently bounded by the quantity and diversity of available demonstrations. In our current experiments, the number of trajectories per task remains limited. As task complexity grows, scaling STG construction while maintaining semantic consistency across a larger state space poses a challenge. 

% \textbf{(2) Offline setting and online generalization.} The current framework assumes a static, offline setting where trajectories are collected in advance. Extending GUITAR to online environments, where the interface state space evolves dynamically, and new states are encountered at inference time, raises open questions about how to incrementally update the STG while preserving diagnostic validity. 

% \textbf{(3) Inference-time integration.} Our guided inference experiments represent a preliminary closed-loop validation. Deploying GUITAR-Guided Inference in practical settings requires addressing efficiency (minimizing latency from STG retrieval and sub-goal generation), robustness (handling incomplete or misleading guidance), and adaptability (updating STG as new behaviors are observed).

% We leave these challenges as directions for future work.

% \section*{References}

\subsubsection*{Acknowledgments and Disclosure of Funding}
This work was supported in part by the National Key R\&D Program of China (2024YFE0215300), in part by the Major Key Project of PCL (PCL2025A12), in part by the Science Fund for Creative Research Groups of the National Natural Science Foundation of China (62521006), in part by the National Natural Science Foundation of China (62406091, 62276077, U23B2055, 62350710797), in part by the Guangdong S\&T Program (2024B0101050003), in part by the Guangdong Basic and Applied Basic Research Foundation (2026A1515011718, 2024A1515011205), in part by the Guangdong Provincial Key Laboratory of Artificial General Intelligence and Large Models (2026B1212010005), and in part by the Shenzhen Science and Technology Program (KQTD20240729102154066). The authors declare no competing interests.

% References follow the acknowledgments in the camera-ready paper. Use unnumbered first-level heading for
% the references. Any choice of citation style is acceptable as long as you are
% consistent. It is permissible to reduce the font size to \verb+small+ (9 point)
% when listing the references.
% Note that the Reference section does not count towards the page limit.
% \medskip

% {
% \small

% [1] Alexander, J.A.\ \& Mozer, M.C.\ (1995) Template-based algorithms for
% connectionist rule extraction. In G.\ Tesauro, D.S.\ Touretzky and T.K.\ Leen
% (eds.), {\it Advances in Neural Information Processing Systems 7},
% pp.\ 609--616. Cambridge, MA: MIT Press.

% [2] Bower, J.M.\ \& Beeman, D.\ (1995) {\it The Book of GENESIS: Exploring
%   Realistic Neural Models with the GEneral NEural SImulation System.}  New York:
% TELOS/Springer--Verlag.

% [3] Hasselmo, M.E., Schnell, E.\ \& Barkai, E.\ (1995) Dynamics of learning and
% recall at excitatory recurrent synapses and cholinergic modulation in rat
% hippocampal region CA3. {\it Journal of Neuroscience} {\bf 15}(7):5249-5262.
% }

%%%%%%%%%%%%%%%%%%%%%%%%%%%%%%%%%%%%%%%%%%%%%%%%%%%%%%%%%%%%
\bibliographystyle{unsrtnat}
\bibliography{neurips_2026}

@article{yang2023auto,
  title={Auto-gpt for online decision making: Benchmarks and additional opinions},
  author={Yang, Hui and Yue, Sifu and He, Yunzhong},
  journal={arXiv preprint arXiv:2306.02224},
  year={2023}
}

@article{yan2023gpt,
  title={Gpt-4v in wonderland: Large multimodal models for zero-shot smartphone gui navigation},
  author={Yan, An and Yang, Zhengyuan and Zhu, Wanrong and Lin, Kevin and Li, Linjie and Wang, Jianfeng and Yang, Jianwei and Zhong, Yiwu and McAuley, Julian and Gao, Jianfeng and others},
  journal={arXiv preprint arXiv:2311.07562},
  year={2023}
}

@article{wu2024atlas,
  title={Os-atlas: A foundation action model for generalist gui agents},
  author={Wu, Zhiyong and Wu, Zhenyu and Xu, Fangzhi and Wang, Yian and Sun, Qiushi and Jia, Chengyou and Cheng, Kanzhi and Ding, Zichen and Chen, Liheng and Liang, Paul Pu and others},
  journal={arXiv preprint arXiv:2410.23218},
  year={2024}
}

@article{qin2025ui,
  title={UI-TARS: Pioneering Automated GUI Interaction with Native Agents},
  author={Qin, Yujia and Ye, Yining and Fang, Junjie and Wang, Haoming and Liang, Shihao and Tian, Shizuo and Zhang, Junda and Li, Jiahao and Li, Yunxin and Huang, Shijue and others},
  journal={arXiv preprint arXiv:2501.12326},
  year={2025}
}

@article{ye2025mobile,
  title={Mobile-agent-v3: Fundamental agents for gui automation},
  author={Ye, Jiabo and Zhang, Xi and Xu, Haiyang and Liu, Haowei and Wang, Junyang and Zhu, Zhaoqing and Zheng, Ziwei and Gao, Feiyu and Cao, Junjie and Lu, Zhengxi and others},
  journal={arXiv preprint arXiv:2508.15144},
  year={2025}
}

@inproceedings{zhang2024dynamic,
  title={Dynamic Planning for LLM-based Graphical User Interface Automation},
  author={Zhang, Shaoqing and Zhang, Zhuosheng and Chen, Kehai and Ma, Xinbei and Yang, Muyun and Zhao, Tiejun and Zhang, Min},
  booktitle={Findings of the Association for Computational Linguistics: EMNLP 2024},
  pages={1304--1320},
  year={2024}
}

@inproceedings{zhou2024webarena,
  title={WEBARENA: A REALISTIC WEB ENVIRONMENT FOR BUILDING AUTONOMOUS AGENTS},
  author={Zhou, Shuyan and Xu, Frank F and Zhu, Hao and Zhou, Xuhui and Lo, Robert and Sridhar, Abishek and Cheng, Xianyi and Ou, Tianyue and Bisk, Yonatan and Fried, Daniel and others},
  booktitle={12th International Conference on Learning Representations, ICLR 2024},
  year={2024}
}

@article{rawles2024androidworld,
  title={Androidworld: A dynamic benchmarking environment for autonomous agents},
  author={Rawles, Christopher and Clinckemaillie, Sarah and Chang, Yifan and Waltz, Jonathan and Lau, Gabrielle and Fair, Marybeth and Li, Alice and Bishop, William and Li, Wei and Campbell-Ajala, Folawiyo and others},
  journal={arXiv preprint arXiv:2405.14573},
  year={2024}
}

@article{xie2024osworld,
  title={Osworld: Benchmarking multimodal agents for open-ended tasks in real computer environments},
  author={Xie, Tianbao and Zhang, Danyang and Chen, Jixuan and Li, Xiaochuan and Zhao, Siheng and Cao, Ruisheng and Hua, Toh J and Cheng, Zhoujun and Shin, Dongchan and Lei, Fangyu and others},
  journal={Advances in Neural Information Processing Systems},
  volume={37},
  pages={52040--52094},
  year={2024}
}

@article{li2025screenspot,
  title={Screenspot-pro: Gui grounding for professional high-resolution computer use},
  author={Li, Kaixin and Meng, Ziyang and Lin, Hongzhan and Luo, Ziyang and Tian, Yuchen and Ma, Jing and Huang, Zhiyong and Chua, Tat-Seng},
  journal={arXiv preprint arXiv:2504.07981},
  year={2025}
}

@inproceedings{bai2021uibert,
  title={UIBert: Learning Generic Multimodal Representations for UI Understanding},
  author={Bai, Chongyang and Zang, Xiaoxue and Xu, Ying and Sunkara, Srinivas and Rastogi, Abhinav and Chen, Jindong and y Arcas, Blaise Ag{\"u}era},
  booktitle={IJCAI},
  year={2021}
}

@inproceedings{deka2017rico,
  title={Rico: A mobile app dataset for building data-driven design applications},
  author={Deka, Biplab and Huang, Zifeng and Franzen, Chad and Hibschman, Joshua and Afergan, Daniel and Li, Yang and Nichols, Jeffrey and Kumar, Ranjitha},
  booktitle={Proceedings of the 30th annual ACM symposium on user interface software and technology},
  pages={845--854},
  year={2017}
}

@inproceedings{cheng2024seeclick,
  title={SeeClick: Harnessing GUI Grounding for Advanced Visual GUI Agents},
  author={Cheng, Kanzhi and Sun, Qiushi and Chu, Yougang and Xu, Fangzhi and YanTao, Li and Zhang, Jianbing and Wu, Zhiyong},
  booktitle={Proceedings of the 62nd Annual Meeting of the Association for Computational Linguistics (Volume 1: Long Papers)},
  pages={9313--9332},
  year={2024}
}

@article{rawles2023androidinthewild,
  title={Androidinthewild: A large-scale dataset for android device control},
  author={Rawles, Christopher and Li, Alice and Rodriguez, Daniel and Riva, Oriana and Lillicrap, Timothy},
  journal={Advances in Neural Information Processing Systems},
  volume={36},
  pages={59708--59728},
  year={2023}
}

@article{li2024effects,
  title={On the effects of data scale on ui control agents},
  author={Li, Wei and Bishop, William E and Li, Alice and Rawles, Christopher and Campbell-Ajala, Folawiyo and Tyamagundlu, Divya and Riva, Oriana},
  journal={Advances in Neural Information Processing Systems},
  volume={37},
  pages={92130--92154},
  year={2024}
}

@article{lu2024gui,
  title={Gui odyssey: A comprehensive dataset for cross-app gui navigation on mobile devices},
  author={Lu, Quanfeng and Shao, Wenqi and Liu, Zitao and Meng, Fanqing and Li, Boxuan and Chen, Botong and Huang, Siyuan and Zhang, Kaipeng and Qiao, Yu and Luo, Ping},
  journal={arXiv preprint arXiv:2406.08451},
  year={2024}
}

@article{zheng2025naturegaia,
  title={NatureGAIA: Pushing the Frontiers of GUI Agents with a Challenging Benchmark and High-Quality Trajectory Dataset},
  author={Zheng, Zihan and Cui, Tianle and Xie, Chuwen and Zhang, Jiahui and Pan, Jiahui and He, Lewei and Chen, Qianglong},
  journal={arXiv preprint arXiv:2508.01330},
  year={2025}
}

@article{zhang2025xbound,
  title={Xbound: Exploring the capability boundaries of device-control agents through trajectory tree exploration},
  author={Zhang, Shaoqing and Chen, Kehai and Zhang, Zhuosheng and Li, Rumei and Weng, Rongxiang and Xiang, Yang and Nie, Liqiang and Zhang, Min},
  journal={arXiv e-prints},
  pages={arXiv--2505},
  year={2025}
}

@inproceedings{zhang2024you,
  title={You Only Look at Screens: Multimodal Chain-of-Action Agents},
  author={Zhang, Zhuosheng and Zhang, Aston},
  booktitle={Findings of the Association for Computational Linguistics ACL 2024},
  pages={3132--3149},
  year={2024}
}

@inproceedings{hong2024cogagent,
  title={Cogagent: A visual language model for gui agents},
  author={Hong, Wenyi and Wang, Weihan and Lv, Qingsong and Xu, Jiazheng and Yu, Wenmeng and Ji, Junhui and Wang, Yan and Wang, Zihan and Dong, Yuxiao and Ding, Ming and others},
  booktitle={Proceedings of the IEEE/CVF Conference on Computer Vision and Pattern Recognition},
  pages={14281--14290},
  year={2024}
}

@article{qian2025webgrapheval,
  title={WebGraphEval: Multi-Turn Trajectory Evaluation for Web Agents using Graph Representation},
  author={Qian, Yaoyao and Wang, Yuanli and Zhang, Jinda and Zong, Yun and Chen, Meixu and Zhou, Hanhan and Huang, Jindan and Zeng, Yifan and Hu, Xinyu and Song, Chan Hee and others},
  journal={arXiv preprint arXiv:2510.19205},
  year={2025}
}

@article{memon2007event,
  title={An event-flow model of GUI-based applications for testing},
  author={Memon, Atif M},
  journal={Software testing, verification and reliability},
  volume={17},
  number={3},
  pages={137--157},
  year={2007},
  publisher={Wiley Online Library}
}

@inproceedings{memon2003gui,
  title={GUI ripping: reverse engineering of graphical user interfaces for testing.},
  author={Memon, Atif M and Banerjee, Ishan and Nagarajan, Adithya},
  booktitle={Wcre},
  volume={3},
  pages={260},
  year={2003}
}

@inproceedings{su2017guided,
  title={Guided, stochastic model-based GUI testing of Android apps},
  author={Su, Ting and Meng, Guozhu and Chen, Yuting and Wu, Ke and Yang, Weiming and Yao, Yao and Pu, Geguang and Liu, Yang and Su, Zhendong},
  booktitle={Proceedings of the 2017 11th joint meeting on foundations of software engineering},
  pages={245--256},
  year={2017}
}

@inproceedings{li2021screen2vec,
  title={Screen2vec: Semantic embedding of gui screens and gui components},
  author={Li, Toby Jia-Jun and Popowski, Lindsay and Mitchell, Tom and Myers, Brad A},
  booktitle={Proceedings of the 2021 CHI Conference on Human Factors in Computing Systems},
  pages={1--15},
  year={2021}
}

@inproceedings{chen2025pg,
  title={PG-Agent: An Agent Powered by Page Graph},
  author={Chen, Weizhi and Wang, Ziwei and Yang, Leyang and Zhou, Sheng and Tang, Xiaoxuan and Bu, Jiajun and Li, Yong and Jiang, Wei},
  booktitle={Proceedings of the 33rd ACM International Conference on Multimedia},
  pages={6878--6887},
  year={2025}
}

@article{gao2026guitester,
  title={GUITester: Enabling GUI Agents for Exploratory Defect Discovery},
  author={Gao, Yifei and Wu, Jiang and Chen, Xiaoyi and Yang, Yifan and Cui, Zhe and Ma, Tianyi and Zhang, Jiaming and Sang, Jitao},
  journal={arXiv preprint arXiv:2601.04500},
  year={2026}
}

@article{zhai2026guide,
  title={GUIDE: Interpretable GUI Agent Evaluation via Hierarchical Diagnosis},
  author={Zhai, Yuwen and Li, Runze and Wang, Liang and Shi, Nian and Xu, Liwu and Zhang, Wei and Lin, Ran and Xu, Bo and Cui, Benlei},
  journal={arXiv preprint arXiv:2604.04399},
  year={2026}
}

@inproceedings{yang2026probench,
  title={Probench: Benchmarking gui agents with accurate process information},
  author={Yang, Leyang and Wang, Ziwei and Tang, Xiaoxuan and Zhou, Sheng and Chen, Dajun and Jiang, Wei and Li, Yong},
  booktitle={Proceedings of the AAAI Conference on Artificial Intelligence},
  volume={40},
  pages={27547--27555},
  year={2026}
}

@article{deng2023mind2web,
  title={Mind2web: Towards a generalist agent for the web},
  author={Deng, Xiang and Gu, Yu and Zheng, Boyuan and Chen, Shijie and Stevens, Sam and Wang, Boshi and Sun, Huan and Su, Yu},
  journal={Advances in Neural Information Processing Systems},
  volume={36},
  pages={28091--28114},
  year={2023}
}

@article{xu2024aguvis,
  title={Aguvis: Unified pure vision agents for autonomous gui interaction},
  author={Xu, Yiheng and Wang, Zekun and Wang, Junli and Lu, Dunjie and Xie, Tianbao and Saha, Amrita and Sahoo, Doyen and Yu, Tao and Xiong, Caiming},
  journal={arXiv preprint arXiv:2412.04454},
  year={2024}
}

@misc{bai2025qwen25vltechnicalreport,
      title={Qwen2.5-VL Technical Report}, 
      author={Shuai Bai and Keqin Chen and Xuejing Liu and Jialin Wang and Wenbin Ge and Sibo Song and Kai Dang and Peng Wang and Shijie Wang and Jun Tang and Humen Zhong and Yuanzhi Zhu and Mingkun Yang and Zhaohai Li and Jianqiang Wan and Pengfei Wang and Wei Ding and Zheren Fu and Yiheng Xu and Jiabo Ye and Xi Zhang and Tianbao Xie and Zesen Cheng and Hang Zhang and Zhibo Yang and Haiyang Xu and Junyang Lin},
      year={2025},
      eprint={2502.13923},
      archivePrefix={arXiv},
      primaryClass={cs.CV},
      url={https://arxiv.org/abs/2502.13923}, 
}

@article{team2026ui,
  title={Ui-venus-1.5 technical report},
  author={Team, Venus and Gao, Changlong and Gu, Zhangxuan and Liu, Yulin and Qiu, Xinyu and Shen, Shuheng and Wen, Yue and Xia, Tianyu and Xu, Zhenyu and Zeng, Zhengwen and others},
  journal={arXiv preprint arXiv:2602.09082},
  year={2026}
}

@article{bai2025qwen3,
  title={Qwen3-vl technical report},
  author={Bai, Shuai and Cai, Yuxuan and Chen, Ruizhe and Chen, Keqin and Chen, Xionghui and Cheng, Zesen and Deng, Lianghao and Ding, Wei and Gao, Chang and Ge, Chunjiang and others},
  journal={arXiv preprint arXiv:2511.21631},
  year={2025}
}

@article{hong2025glm,
  title={Glm-4.5 v and glm-4.1 v-thinking: Towards versatile multimodal reasoning with scalable reinforcement learning},
  author={Hong, Wenyi and Yu, Wenmeng and Gu, Xiaotao and Wang, Guo and Gan, Guobing and Tang, Haomiao and Cheng, Jiale and Qi, Ji and Ji, Junhui and Pan, Lihang and others},
  journal={arXiv preprint arXiv:2507.01006},
  year={2025}
}

@article{tang2025survey,
  title={A survey on (m) llm-based gui agents},
  author={Tang, Fei and Xu, Haolei and Zhang, Hang and Chen, Siqi and Wu, Xingyu and Shen, Yongliang and Zhang, Wenqi and Hou, Guiyang and Tan, Zeqi and Yan, Yuchen and others},
  journal={arXiv preprint arXiv:2504.13865},
  year={2025}
}

@inproceedings{ji2026thinking,
  title={Thinking with map: Reinforced parallel map-augmented agent for geolocalization},
  author={Ji, Yuxiang and Wang, Yong and Ma, Ziyu and Hu, Yiming and Huang, Hailang and Hu, Xuecai and Chen, Guanhua and Wu, Liaoni and Chu, Xiangxiang},
  booktitle={Findings of the Association for Computational Linguistics: ACL 2026},
  pages={17991--18004},
  year={2026}
}

@inproceedings{deng2025retail,
  title={Retail: Towards real-world travel planning for large language models},
  author={Deng, Bin and Feng, Yizhe and Liu, Zeming and Wei, Qing and Zhu, Xiangrong and Chen, Shuai and Guo, Yuanfang and Wang, Yunhong},
  booktitle={Proceedings of the 2025 Conference on Empirical Methods in Natural Language Processing},
  pages={14881--14913},
  year={2025}
}

\newpage

% !TeX root = ../neurips_2026.tex
\appendix

\section*{Limitations}\label{sec:lmitations}

While GUITAR demonstrates promising diagnostic capabilities, several limitations remain.

\textbf{(1) Dependency on STG Construction Quality.}
GUITAR relies on the quality of its functional state abstractions. Although the five-stage construction pipeline is fully automatic, assignment errors may still affect the identified bottlenecks. Our comparison between automatic and manually verified STGs shows that the principal failure structure is largely preserved, but improving automatic abstraction remains important for more complex and dynamic interfaces.

\textbf{(2) Dataset Scale and Generalization Scope.}
We evaluate GUITAR on six primary tasks from AndroidControl and Mind2Web, selected according to trajectory availability. The trajectory-held-out evaluation demonstrates generalization to unseen trajectories within the same task environments, but does not establish transfer to unseen applications, websites, or interaction domains. Evaluating GUITAR on larger and more diverse environments is therefore an important direction for future work.

\textbf{(3) Offline Construction and Static STGs.}
GUITAR currently constructs STGs offline from a fixed set of reference trajectories. Consequently, rare states absent from the reference data may not be represented, and changes to the target application may require graph reconstruction. Online or incremental STG updates would improve scalability and adaptability.

\textbf{(4) Guided Inference as a Proof of Concept.}
The guided-inference experiment is intended to validate the actionable value of the identified bottlenecks rather than provide a fully optimized inference system. Targeted guidance improves SR by 2.8 percentage points in the primary setting and retains a gain of 1.88 percentage points under trajectory-held-out evaluation, but gains vary across agents and tasks. Better retrieval, sub-goal generation, and agent-specific guidance remain open directions.

\section{Construction and Evaluation Details}
\label{sec:annotation_details}

\subsection{STG Construction Details}\label{sec:stg_construction}

\begin{figure}[h]
    \centering
    \includegraphics[width=1\linewidth]{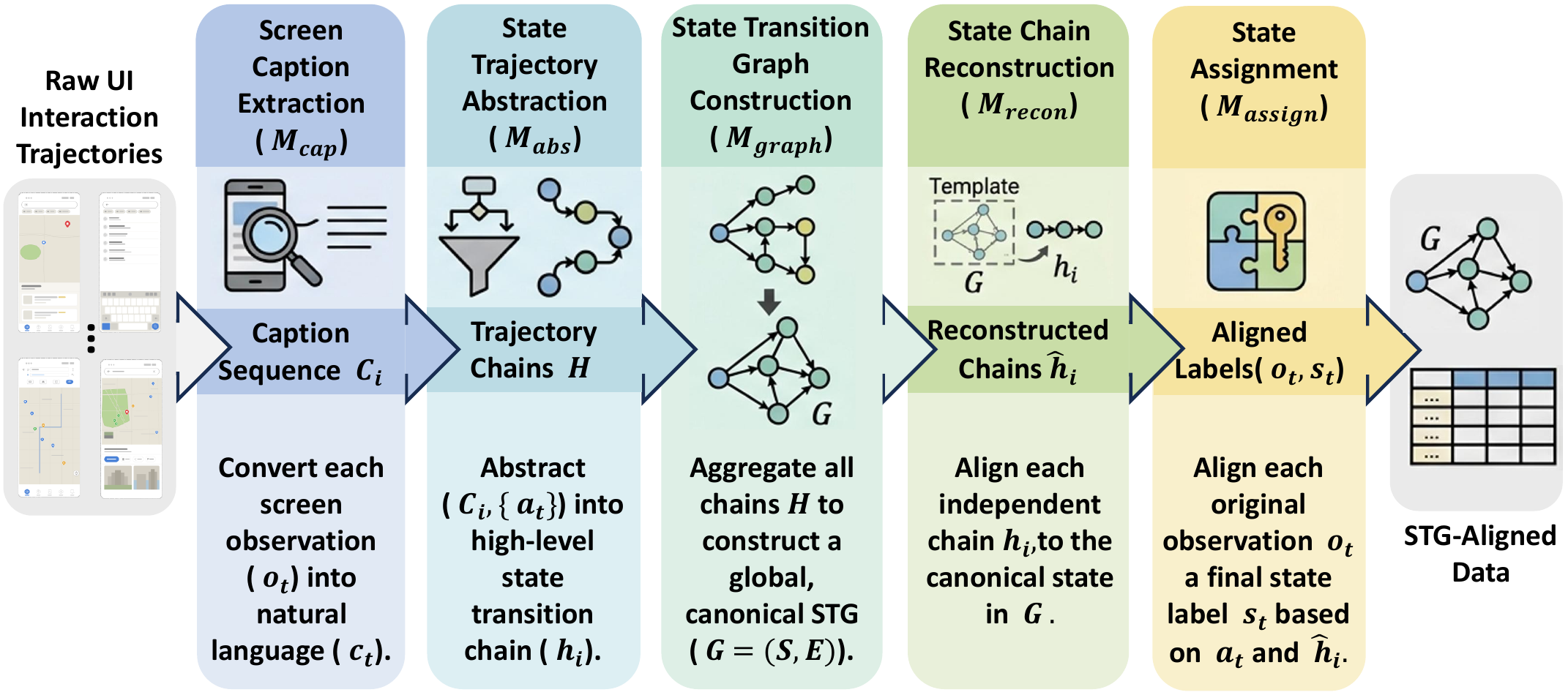}
    \caption{The fully automatic STG construction pipeline. It transforms raw UI interaction trajectories into a State Transition Graph through five VLM-based stages: (1) extracting semantic captions via $\mathcal{M}_{cap}$, (2) abstracting trajectories into state chains via $\mathcal{M}_{abs}$, (3) aggregating chains into a global STG via $\mathcal{M}_{graph}$, (4) reconstructing chains for semantic consistency via $\mathcal{M}_{recon}$, and (5) performing fine-grained screenshot-to-state assignment via $\mathcal{M}_{assign}$. Manual verification, when used, is a separate post-hoc step and is not part of this pipeline.}
    \label{fig:build_graph_pipeline}
    % \vspace{-1.5ex}
\end{figure}

We construct the canonical STG topology from successfully executed trajectories so that its state and transition vocabulary is grounded in valid task workflows rather than erroneous actions. 
As illustrated in Figure~\ref{fig:build_graph_pipeline}, the construction pipeline progressively transforms low-level screen observations into a structured, semantically consistent graph. All five stages are performed automatically using Qwen2.5-VL-72B. The output of Stage~5 is therefore a fully automatic STG; the post-hoc human verification described below is used only to obtain the manually verified variant employed in our primary diagnostic analysis.
% An example is summarized in Algorithm~\ref{alg:construct_stg_example}.

\paragraph{Notation.}
Let $\mathcal{T} = \{\tau_1, \dots, \tau_N\}$ denote a set of trajectories, where each trajectory $\tau_i = (g_i, \{(o_t, a_t)\}_{t=1}^{T_i})$ contains a task goal $g_i$, screen observations $o_t$, and corresponding user actions $a_t$. 
We use $c_t$ to denote the natural language caption of $o_t$, and $\mathcal{C}_i = \{c_t\}$ the caption sequence of trajectory $\tau_i$. 
Each trajectory is abstracted into a state transition chain $h_i$, and we denote the set of all chains as $\mathcal{H} = \{h_i\}$. 
The final STG is defined as $\mathcal{G} = (\mathcal{S}, \mathcal{E})$, where $\mathcal{S}$ is the set of states and $\mathcal{E}$ is the set of transitions. 
After graph construction, each trajectory is aligned to the graph as $\hat{h}_i$, and each observation is assigned a state label $s_t \in \mathcal{S}$. 
We denote different functional modules of the VLMs as $\mathcal{M}_{cap}$, $\mathcal{M}_{abs}$, $\mathcal{M}_{graph}$, $\mathcal{M}_{recon}$, and $\mathcal{M}_{assign}$, corresponding to different stages in the pipeline.

\paragraph{Operational Criterion for Functional Equivalence.}
State identity is determined by a screen's role in the task trajectory rather than by visual appearance alone. Two observations are assigned to the same state when, conditioned on the task goal and trajectory context, they support the same task-relevant operation and occupy the same functional position in the workflow. Conversely, visually similar observations are kept separate when they require different next decisions or lead to different downstream transitions. The pipeline operationalizes this criterion using the screen caption, the action executed between consecutive observations, and the surrounding state-transition context. Thus, visually different pages are merged only when their action-conditioned trajectory roles agree.

\paragraph{Stage 1: Screen Caption Extraction.}
We first convert each screen observation into a natural language description. 
Specifically, for each $o_t$, we obtain a caption $c_t = \mathcal{M}_{cap}(o_t)$, producing a caption sequence $\mathcal{C}_i$ for each trajectory. 
These captions provide a unified semantic representation of UI states, bridging visual content and language-based reasoning. This stage consumes the raw screenshot but does not yet decide state identity.

\paragraph{Stage 2: State Trajectory Abstraction.}
Given the task goal $g_i$, caption sequence $\mathcal{C}_i$, and action sequence $\{a_t\}$, we abstract each trajectory into a state transition chain:
\[
h_i = \mathcal{M}_{abs}(g_i, \mathcal{C}_i, \{a_t\})
\]
where each node represents a high-level UI state and each edge corresponds to a user action. 
This step compresses low-level interactions into semantically meaningful transitions. In particular, the action between consecutive observations provides behavioral evidence for the functional role of each state.

\paragraph{Stage 3: State Transition Graph Construction.}
We aggregate all trajectory-level chains $\mathcal{H}$ and construct a global STG:
\[
\mathcal{G} = \mathcal{M}_{graph}(\mathcal{H})
\]
This process merges states with equivalent action-conditioned trajectory roles across trajectories and resolves redundant descriptions, resulting in a unified graph structure. No raw images are used in this aggregation stage.

\paragraph{Stage 4: State Chain Reconstruction.}
Since trajectory-level abstractions are generated independently, their state representations may be inconsistent. 
To address this, we align each chain $h_i$ with the canonical state space defined by $\mathcal{G}$:
\[
\hat{h}_i = \mathcal{M}_{recon}(h_i, \mathcal{G})
\]
This step enforces consistency across trajectories by mapping states to shared semantic identities.

\paragraph{Stage 5: State Assignment.}
We assign each observation $o_t$ to a state in $\mathcal{G}$ by considering the task goal, its caption, the preceding action history $a_{<t}$, the next action $a_t$, and the reconstructed trajectory context:
\[
s_t = \mathcal{M}_{assign}(o_t, c_t, g_i, a_{<t}, a_t, \hat{h}_i)
\]
This produces a fine-grained alignment between raw observations and abstract states. 
Stages~1 and~5 are the only construction stages that directly consume raw screenshots; Stages~2--4 operate on textual state/action representations.

\paragraph{Post-hoc Manual Verification.}
For the manually verified STG variant used in our primary diagnostic analysis, annotators verify the screenshot-to-state assignments produced by Stage~5 using the reconstructed trajectory chains from Stage~4 as context. When an assignment is incorrect, the annotator reassigns the screenshot to an existing state with the appropriate functional role. This step corrects only the final observation-to-state mapping; annotators do not define state semantics or construct the graph topology. It is also not required to run the five-stage pipeline: the unverified output of Stage~5 is the fully automatic STG used in our automatic-pipeline evaluations. Table~\ref{tab:correction_rate} reports the fraction of assignments changed during this post-hoc verification.

\begin{figure}[t]
    \centering
    \includegraphics[width=1\linewidth]{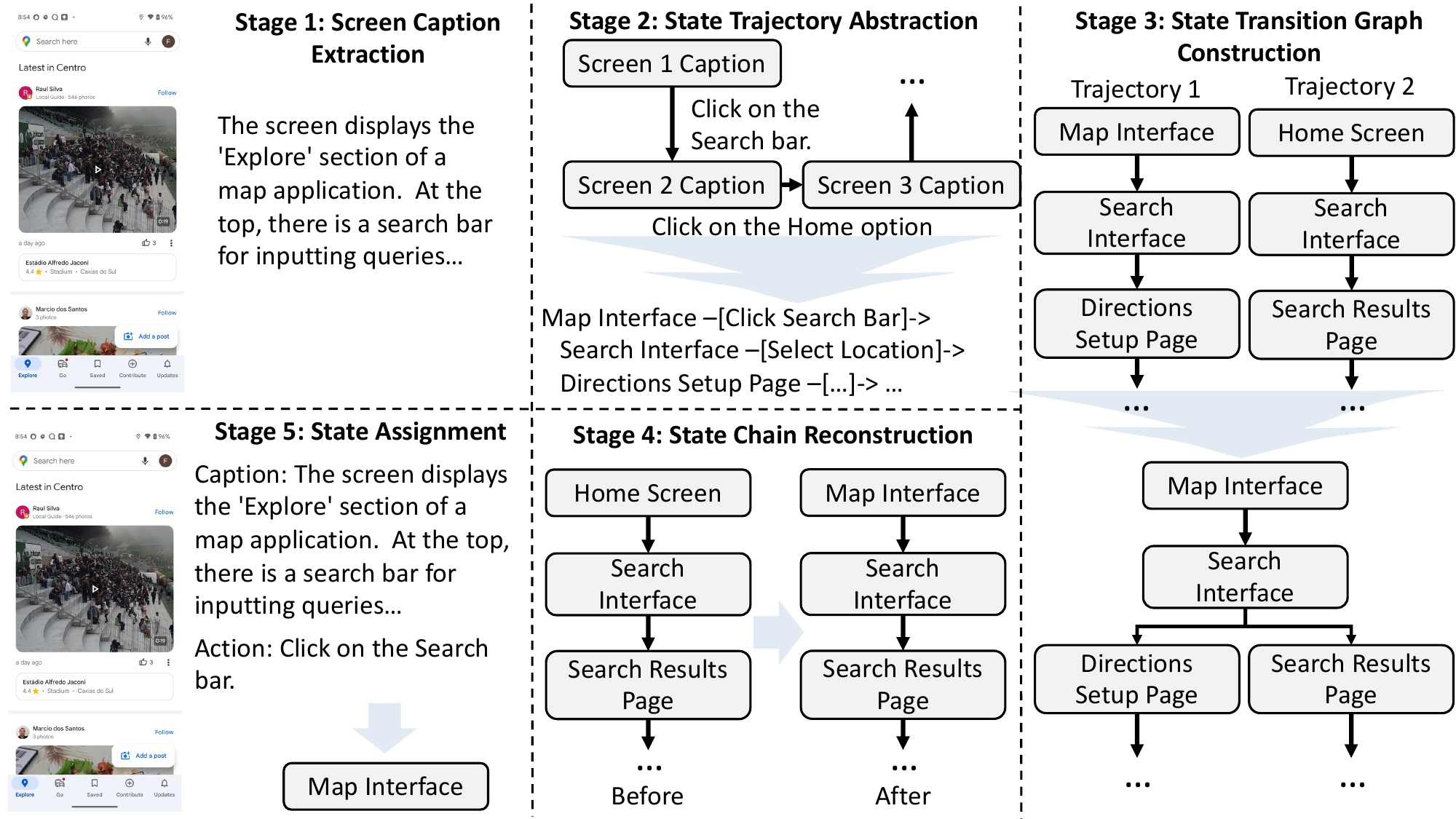}
    \caption{An illustrative example of STG construction. Stage 1: extract screen captions from raw screenshots. Stage 2: abstract state trajectories from captions and actions. Stage 3: merge state trajectories into a unified STG. Stage 4: refine state trajectories guided by the constructed STG. Stage 5: assign each screen to its best-matching state based on the screenshot, caption, and executed action.}
    \label{fig:construct_stg_example}
    % \vspace{-1.5ex}
\end{figure}

\begin{table}[b]
\centering
\caption{Post-hoc screenshot-to-state correction rates across tasks. A correction changes the Stage~5 assignment but does not redefine state semantics or graph topology.}
\label{tab:correction_rate}
\begin{tabular}{llc}
\toprule
\textbf{Platform} & \textbf{Task} & \textbf{Correction Rate (\%)} \\
\midrule
\multirow{3}{*}{Android} & Kitchen Stories & 5.1  \\
                         & Maps           & 25.3 \\
                         & CNN            & 0.0  \\
\midrule
\multirow{3}{*}{Web}     & Booking        & 20.8 \\
                         & Carmax         & 22.7 \\
                         & Kayak          & 23.7 \\
\bottomrule
\end{tabular}
\end{table}

As shown in Table~\ref{tab:correction_rate}, correction rates range from 0.0\% to 25.3\%. CNN and Kitchen Stories require little or no correction, whereas Maps and the three web tasks require more intervention. The latter settings contain richer dynamic content and more varied layouts, which make fine-grained screenshot-to-state assignment more difficult. These rates quantify post-hoc changes to Stage~5 assignments, rather than the fraction of the construction pipeline performed by humans.

\subsection{STG Evaluation Details}\label{sec:stg_evaluation}

\begin{figure}[t]
    \centering
    \includegraphics[width=1\linewidth]{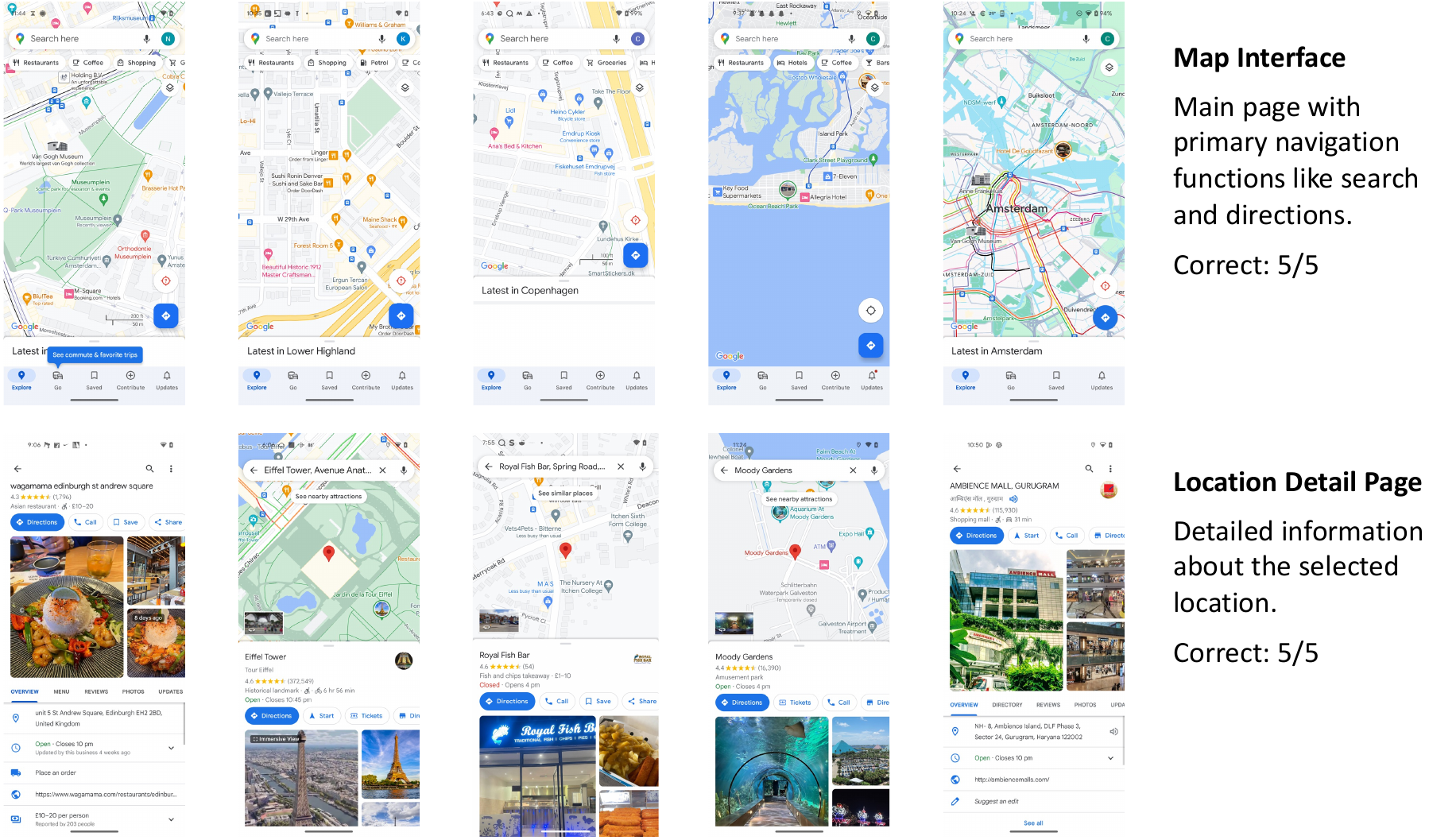}
    \caption{Human evaluation of STG state classification accuracy. Annotators are shown screenshots and asked whether each matches its assigned state label and description. Numbers indicate the count of confirmed matches per state.}
    \label{fig:human_annotation1}
    % \vspace{-1.5ex}
\end{figure}

We design a human evaluation protocol to assess the quality of the constructed STG, focusing on the consistency between screen observations and their assigned state labels. 
For each state $s \in \mathcal{S}$, we randomly sample up to five screen observations associated with that state.\footnote{If fewer than five observations are available, all observations are used.} 
As shown in Figure~\ref{fig:human_annotation1}, annotators are presented with the state description and the sampled images, and are asked to judge whether each image is consistent with the semantic meaning of the state. 
Each judgment is binary, indicating consistency or inconsistency.

To measure inter-annotator agreement, we compute Gwet's AC1 on the resulting binary state-level labels across annotators. 
We adopt AC1 instead of Cohen's kappa due to its robustness to prevalence and marginal distribution issues, which are common in skewed labeling scenarios. 
A higher AC1 score indicates stronger agreement among annotators on the overall quality of the STG.
To obtain a state-level quality label, we aggregate the image-level judgments. 
Specifically, a state is considered valid if the proportion of images judged as consistent exceeds a predefined threshold $\delta$. In our experiments, we set $\delta = 0.8$. 

\subsection{Similarity-Based Abstraction and Evaluation}\label{sec:abstraction_evaluation}

\subsubsection{Low-Level Similarity Clustering}

We employ ImageHash to compute perceptual similarity for low-level state clustering.
A screenshot is merged into an existing state if either of the following conditions is satisfied:

\begin{itemize}
    \item The Hamming distance between the image's hash and the hashes stored within the state is less than 25; or
    \item The cosine similarity between the action text embedding of the current screenshot and the action text vector maintained within the state exceeds 0.75.
\end{itemize}

Upon merging, the action text vector of the target state is updated accordingly to reflect the newly incorporated action.

\subsubsection{Abstraction Evaluation}

To quantitatively evaluate the quality of the constructed abstraction, we assess both visual and behavioral consistency across states and transitions.

\paragraph{Models.}
We adopt the following pre-trained models for similarity computation:
\begin{itemize}
    \item \textbf{Image similarity:} \texttt{clip-vit-base-patch16}
    \item \textbf{Text similarity:} \texttt{all-MiniLM-L6-v2}
\end{itemize}

\paragraph{Intra-State Image Similarity.}
For each state, we compute the pairwise image similarity among all screenshots contained within it using the CLIP model. We report the following statistics:
\begin{itemize}
    \item Mean and standard deviation of pairwise similarity scores;
    \item Ratio: the proportion of image pairs whose similarity score exceeds the threshold of 0.70.
\end{itemize}

\paragraph{Transition-Level Action Similarity.}
To evaluate behavioral consistency across transitions, we compute text similarity between action embeddings at the transition level:
\begin{itemize}
    \item \textbf{Intra-transition similarity:} average similarity among actions belonging to the same transition;
    \item \textbf{Inter-transition similarity:} average similarity among actions belonging to different transitions.
\end{itemize}

To assess the statistical significance of the observed intra-transition cohesion, we conduct a Permutation Test:
\begin{enumerate}
    \item All action embeddings are randomly shuffled and reassigned into groups matching the original transition group sizes.
    \item The mean intra-group similarity is computed for each permutation.
    \item This procedure is repeated 1000 times to construct a null distribution, against which the observed similarity is compared.
\end{enumerate}

\section{Complete Experimental Results}
\label{sec:full_results}

\begin{figure*}[hb]
    \centering
    \begin{subfigure}[t]{0.95\linewidth}
        \includegraphics[width=\linewidth]{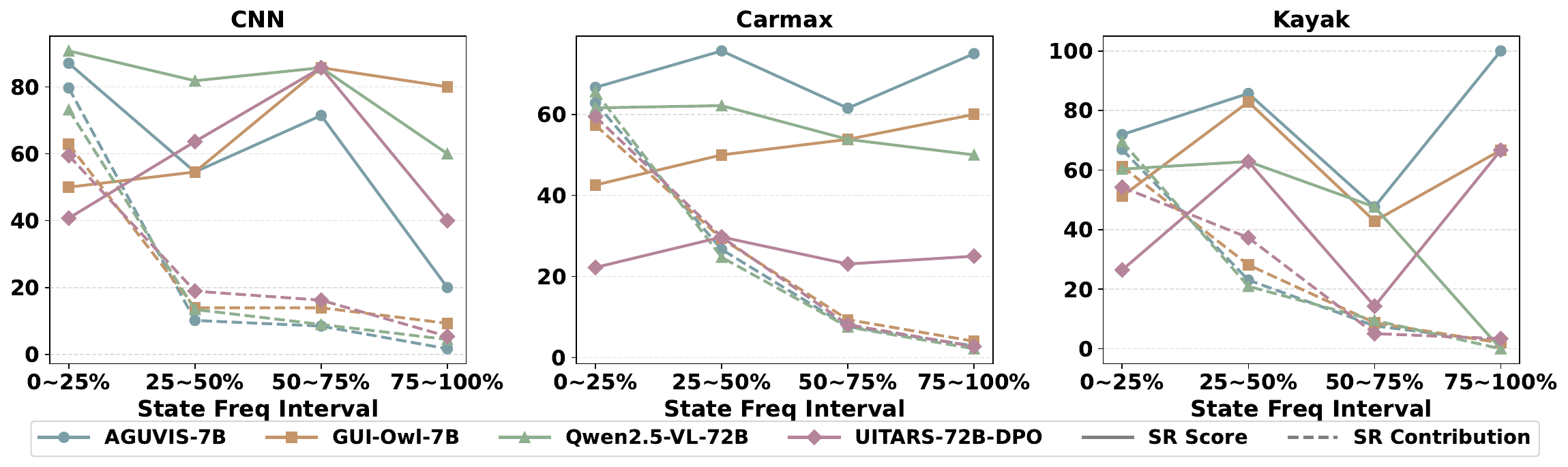}
    \end{subfigure}
    \hfill
    \begin{subfigure}[t]{0.95\linewidth}
        \includegraphics[width=\linewidth]{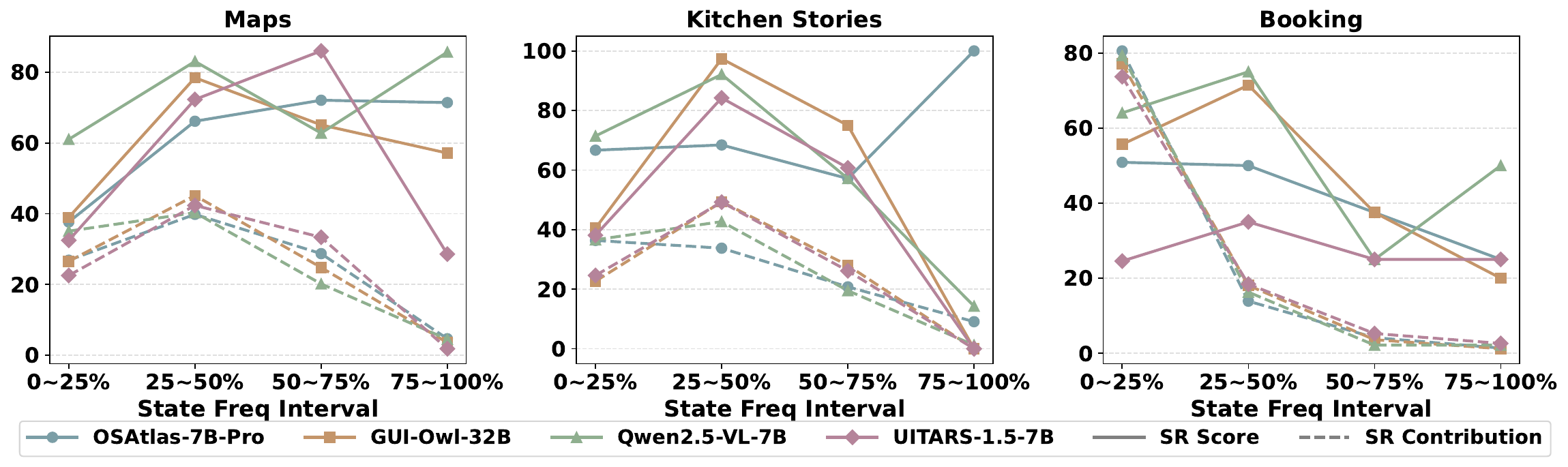}
    \end{subfigure}
    \hfill
    \begin{subfigure}[t]{0.95\linewidth}
        \includegraphics[width=\linewidth]{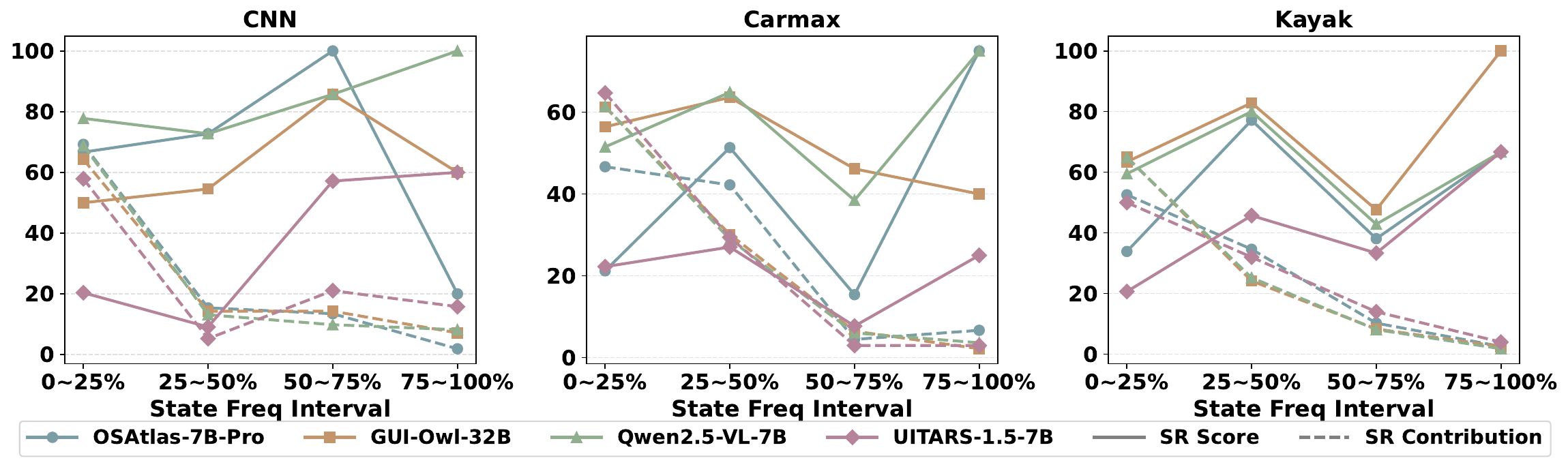}
    \end{subfigure}
    \caption{Analysis of SR contribution across state frequency intervals for 8 models over 6 tasks. Across all settings, low-frequency states contribute marginally to the overall SR regardless of their individual performance, confirming that SR systematically underrepresents agent behavior in the long tail.}
    \label{fig:freq_contribution2}
    % \vspace{-4ex}
\end{figure*}

\subsection{Contribution Imbalance}

GUI interaction naturally follows a long-tailed distribution, where a small number of high-frequency states dominate trajectories while the majority of states are visited rarely.
This structural property has a direct consequence for evaluation: step-level metrics are disproportionately influenced by high-frequency states, leaving agent performance on rare but critical states largely unaccounted for.

Figure~\ref{fig:freq_contribution2} illustrates the SR contribution across state frequency intervals for 8 models over 6 tasks. Across all settings, low-frequency states contribute marginally to the overall SR regardless of their individual performance, confirming that SR systematically underrepresents agent behavior in the long tail.

\subsection{Failure Concentration Across States}

\begin{figure*}[hb]
    \centering
    \begin{subfigure}[b]{0.93\linewidth}
        \includegraphics[width=\linewidth]{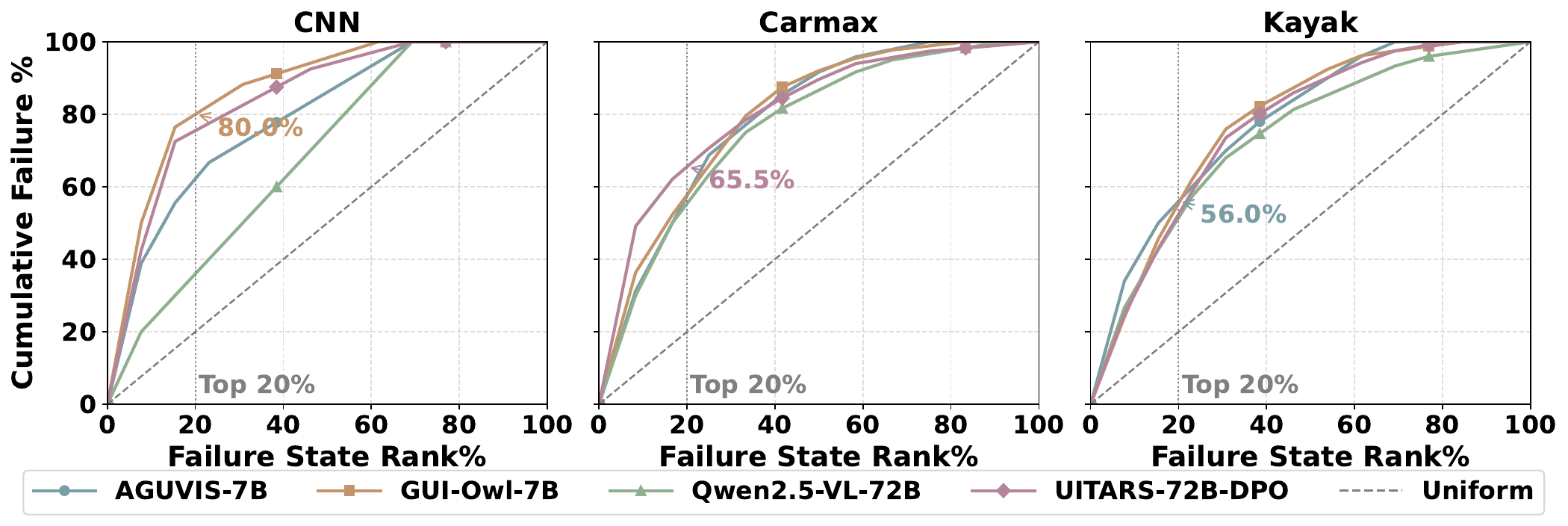}
    \end{subfigure}
    \hfill
    \begin{subfigure}[b]{0.93\linewidth}
        \includegraphics[width=\linewidth]{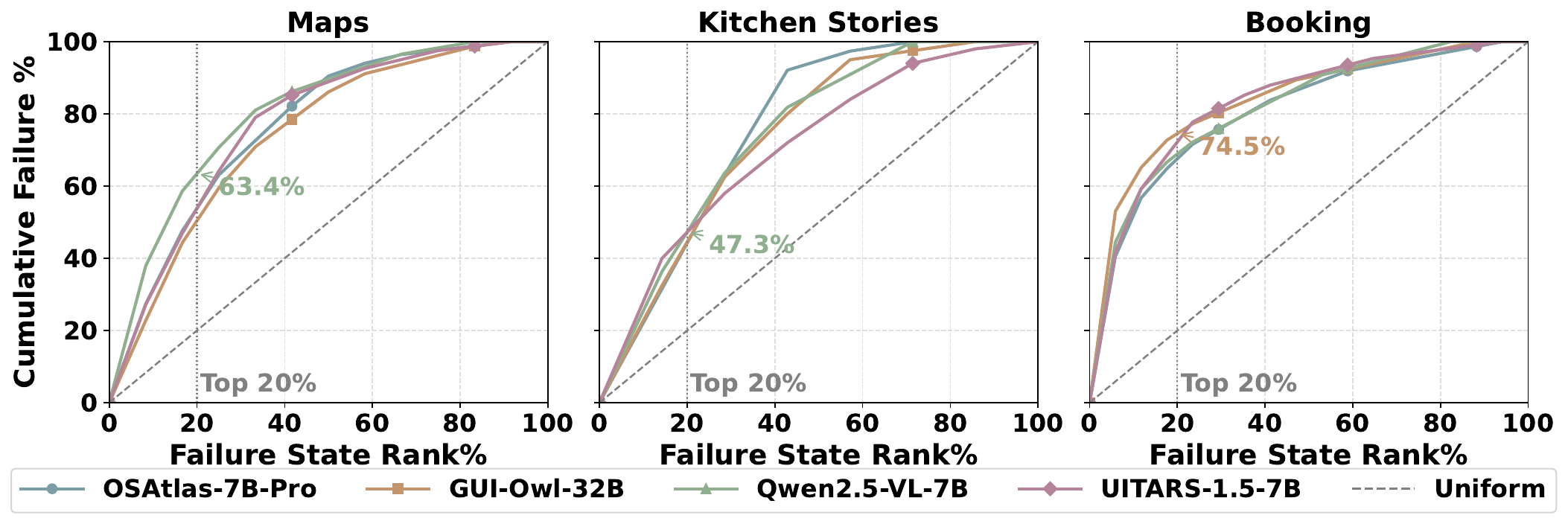}
    \end{subfigure}
    \hfill
    \begin{subfigure}[b]{0.93\linewidth}
        \includegraphics[width=\linewidth]{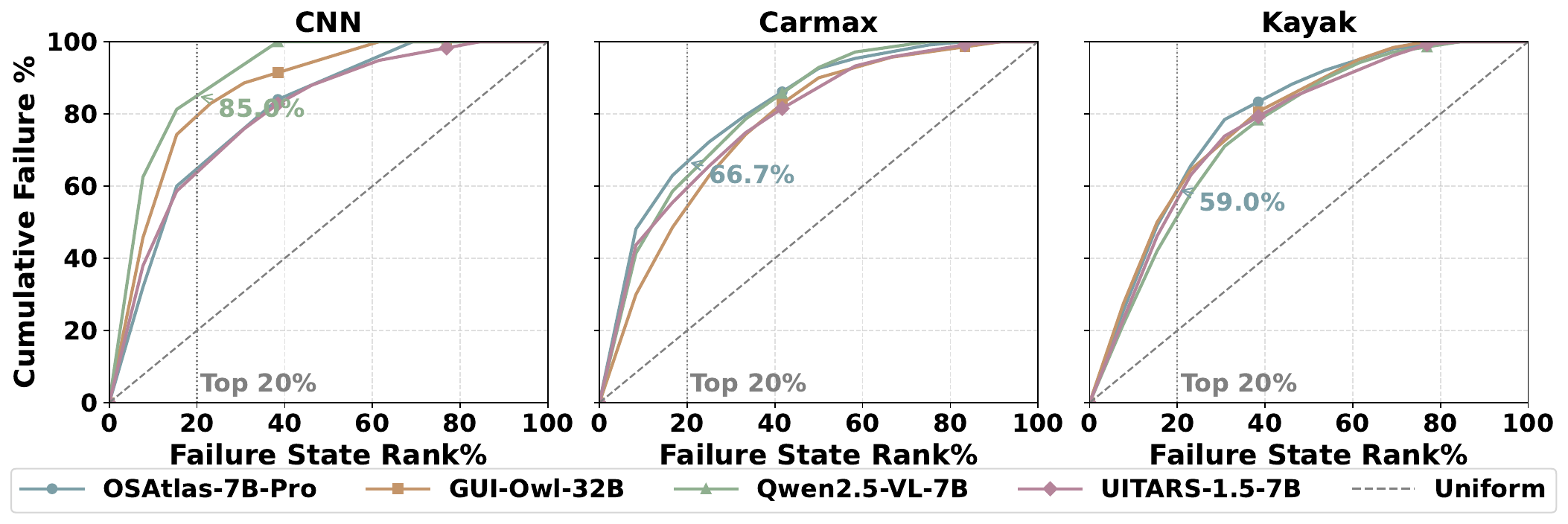}
    \end{subfigure}
    \caption{Cumulative distribution of failures 
    across states ranked by failure frequency for 
    8 models over 6 tasks. The top 20\% of states 
    account for the majority of total errors across 
    all settings, indicating strong and consistent 
    failure concentration.}
    \label{fig:failure_distribution2}
    \vspace{-1ex}
\end{figure*}

We analyze the distribution of failures across states for 8 models over 6 tasks.
As shown in Figure~\ref{fig:failure_distribution2}, the top 20\% of states ranked by failure frequency account for the majority of total errors across all settings, revealing that agent failures are highly concentrated rather than uniformly distributed. 
For instance, this concentration reaches 80\% for GUI-Owl-7B on CNN and 65.5\% for UI-TARS-72B-DPO on Carmax, demonstrating that a small set of bottleneck states drives the majority of failures regardless of model capability.
This concentration suggests that targeted intervention on a small set of bottleneck states is sufficient to address the majority of agent failures.

\subsection{Controlled SFT--DPO Comparison}
\label{sec:sft_dpo_comparison}

To examine whether post-training strategy affects model-specific bottlenecks, we compare UI-TARS-7B-SFT and UI-TARS-7B-DPO while holding the base model and action space fixed. Table~\ref{tab:sft_dpo_comparison} shows that DPO slightly reduces average failure concentration (Gini: $0.555$ vs. $0.589$), while the state rankings remain strongly correlated (mean Spearman $\rho=0.913$). However, the overlap is not complete (mean Top-20\% Jaccard: $0.778$), indicating that post-training redistributes failure mass across states rather than uniformly improving all states.

\begin{table*}[t]
\centering\small
\renewcommand{\arraystretch}{1.05}
\setlength{\tabcolsep}{3.2mm}
\caption{Controlled comparison of failure patterns between UI-TARS-7B-SFT and UI-TARS-7B-DPO. Jaccard measures the overlap between their Top-20\% bottleneck states, and Spearman $\rho$ compares the full state-level failure rankings.}
\label{tab:sft_dpo_comparison}
\begin{tabular}{lcccc}
\toprule
\textbf{Task} & \textbf{Gini (SFT)} & \textbf{Gini (DPO)} & \textbf{Jaccard} & \textbf{Spearman $\rho$} \\
\midrule
Maps            & 0.648 & 0.571 & 0.333 & 0.972 \\
Kitchen Stories & 0.464 & 0.449 & 1.000 & 0.905 \\
CNN             & 0.654 & 0.644 & 1.000 & 0.863 \\
\midrule
\textbf{Mean}   & \textbf{0.589} & \textbf{0.555} & \textbf{0.778} & \textbf{0.913} \\
\bottomrule
\end{tabular}
\end{table*}

The largest compositional shift occurs on Maps: the SFT model's unique Top-20\% bottleneck is the \textit{Route Map Page}, whose inspected failures are mostly discriminative-action errors such as misjudging the map state or selecting an incorrect target. In contrast, the DPO model's unique bottleneck is the \textit{Start Location Search Interface}, where the inspected failures are mostly generative-action errors such as entering an incorrect location. Across the three tasks, DPO reduces the error rate on discriminative actions by $0.015$ on average, more than on generative actions. These results suggest that DPO can shift both bottleneck location and error composition, but they constitute evidence from one controlled model pair rather than a causal conclusion across training paradigms.

\section{GUITAR-Guided Inference}

\subsection{Full Trajectory-Held-out Results}
\label{sec:extended_heldout}

The main paper reports trajectory-held-out guidance on the six primary tasks. We apply the same three-fold protocol to three additional AndroidControl tasks---Arts \& Culture, Artsy, and Pinterest---and summarize the complete nine-task evaluation in Table~\ref{tab:extended_heldout}. In each fold, the fully automatic STG topology and state-level error statistics are constructed from two thirds of the trajectories, while the untouched third is used only for guided-inference evaluation.

\begin{table*}[t]
\centering\small
\renewcommand{\arraystretch}{1.05}
\caption{Full three-fold trajectory-held-out guidance results with fully automatic STGs, summarized by task (Panel A) and agent (Panel B). $N$ is the number of evaluated task--agent pairs, and I/T/D denotes improved/tied/decreased. Overall values average all 57 available pairs.}
\label{tab:extended_heldout}
\begin{minipage}[t]{0.43\textwidth}
\centering
\setlength{\tabcolsep}{1.2mm}
\begin{tabular}{lccc}
\toprule
\multicolumn{4}{c}{\textbf{Panel A: Results by task}} \\
\textbf{Task} & \textbf{$N$} & \textbf{Mean $\Delta$SR (pp)} & \textbf{I/T/D} \\
\midrule
CNN              & 7  & $+0.87$ & $4/3/0$ \\
Kitchen Stories  & 7  & $+3.29$ & $5/2/0$ \\
Maps             & 7  & $+2.98$ & $4/3/0$ \\
Arts \& Culture  & 7  & $+6.04$ & $5/0/2$ \\
Artsy            & 7  & $+4.37$ & $5/0/2$ \\
Pinterest        & 7  & $+5.09$ & $7/0/0$ \\
Booking          & 5  & $+1.20$ & $1/4/0$ \\
Carmax           & 5  & $+1.09$ & $2/2/1$ \\
Kayak            & 5  & $+1.27$ & $1/2/2$ \\
\midrule
\textbf{Overall} & \textbf{57} & \textbf{$+3.09$} & \textbf{$34/16/7$} \\
\bottomrule
\end{tabular}
\end{minipage}%
\hfill
\begin{minipage}[t]{0.53\textwidth}
\centering
\setlength{\tabcolsep}{0.8mm}
\begin{tabular}{lccc}
\toprule
\multicolumn{4}{c}{\textbf{Panel B: Results by agent}} \\
\textbf{Agent} & \textbf{$N$} & \textbf{Mean $\Delta$SR (pp)} & \textbf{I/T/D} \\
\midrule
Aguvis-7B          & 9 & $+1.39$ & $4/3/2$ \\
GUI-Owl-7B         & 9 & $+5.23$ & $7/1/1$ \\
OS-Atlas-7B        & 9 & $+3.51$ & $7/1/1$ \\
Qwen2.5-VL-7B      & 9 & $+0.29$ & $4/3/2$ \\
UI-TARS-1.5-7B     & 9 & $+3.00$ & $3/5/1$ \\
GUI-Owl-32B        & 6 & $+5.38$ & $6/0/0$ \\
UI-TARS-72B-DPO    & 6 & $+3.85$ & $3/3/0$ \\
\midrule
\textbf{Overall}  & \textbf{57} & \textbf{$+3.09$} & \textbf{$34/16/7$} \\
\bottomrule
\end{tabular}
\end{minipage}

\vspace{0.5ex}
\raggedright\footnotesize GUI-Owl-32B and UI-TARS-72B-DPO are evaluated on the six AndroidControl tasks; the other five agents are evaluated on all nine tasks.
\end{table*}

Across the 57 task--agent pairs, guidance improves SR by 3.09 percentage points on average, with 34 improvements, 16 ties, and 7 decreases. The mean gain is positive on every task and for every agent, supporting within-task generalization to unseen trajectories across the evaluated environments.

\paragraph{Held-out State Coverage.}
Across 771 held-out steps, the mean top-1 state similarity is $0.770$, and the out-of-graph rate $P(\mathrm{sim}_{\mathrm{top1}}<\tau_{\mathrm{sim}})$ is $3.89\%$ at $\tau_{\mathrm{sim}}=0.6$. The three fold-specific rates are $6.25\%$, $2.36\%$, and $2.62\%$ (standard deviation: $1.8$ percentage points), indicating that low-confidence state matching is uncommon under the evaluated node libraries.

\subsection{Evaluation on Recent Model Families}
\label{sec:recent_model_guidance}

We further evaluate Qwen3-VL (2B/4B/8B), UI-Venus-1.5 (2B/8B), and GLM-4.6V-Flash under the same guidance protocol on the six primary tasks~\citep{team2026ui,bai2025qwen3,hong2025glm}. Table~\ref{tab:recent_model_guidance} summarizes the results against the corresponding unguided baselines. For Qwen3-VL, guidance yields an average SR gain of 4.00 percentage points over 18 task--model pairs (14 improved, 1 tied, and 3 decreased); the model-wise gains are +4.61 pp, +6.67 pp, and +0.72 pp for 2B, 4B, and 8B, respectively. UI-Venus-1.5 and GLM-4.6V-Flash jointly improve SR by 2.58 percentage points over 18 pairs (12 improved and 6 decreased). Across all six recent models, the mean gain is 3.29 percentage points (26 improved, 1 tied, and 9 decreased), showing a positive average effect without implying uniform gains across every model--task pair.

\begin{table}[t]
\centering\small
\renewcommand{\arraystretch}{1.05}
\setlength{\tabcolsep}{2.5mm}
\caption{Guidance performance ($\Delta$SR, pp) on recent model families over the six primary tasks. The Qwen3-VL row is averaged across its 2B, 4B, and 8B variants; the remaining rows report individual models.}
\label{tab:recent_model_guidance}
\resizebox{\textwidth}{!}{%
\begin{tabular}{lccccccc}
\toprule
\textbf{Model} & \textbf{Maps} & \textbf{Kitchen} & \textbf{CNN} & \textbf{Booking} & \textbf{Carmax} & \textbf{Kayak} & \textbf{Mean} \\
\midrule
Qwen3-VL (2B/4B/8B avg.) & $+9.05$ & $+7.75$ & $+3.79$ & $+2.85$ & $+0.17$ & $+0.40$ & \textbf{$+4.00$} \\
UI-Venus-1.5-2B          & $+3.11$ & $+2.32$ & $+7.95$ & $+1.67$ & $+3.04$ & $-3.32$ & \textbf{$+2.46$} \\
UI-Venus-1.5-8B          & $-0.38$ & $+1.44$ & $+9.97$ & $+4.72$ & $+4.30$ & $-1.52$ & \textbf{$+3.09$} \\
GLM-4.6V-Flash           & $-0.61$ & $+3.60$ & $+11.49$ & $+2.62$ & $-1.69$ & $-2.28$ & \textbf{$+2.19$} \\
\bottomrule
\end{tabular}%
}
\end{table}

\subsection{GUITAR-Guided Inference Implementation Details}\label{sec:guitar_inference}

GUITAR-Guided Inference operates in two stages: state matching and STG-guided sub-goal generation. 
State matching is performed by retrieving the top-$K$ ($K=3$) most similar states from the STG. 
For each candidate state, a fused similarity score is computed by combining textual and visual signals: $s = 0.5 \times s_{\text{text}} + 0.5 \times s_{\text{visual}}$, where $s_{\text{text}}$ is computed using \texttt{all-MiniLM-L6-v2} over the full execution history and $s_{\text{visual}}$ is computed using \texttt{clip-vit-base-patch16} over the current screenshot. 
Both signals are computed simultaneously and fused into a score for state retrieval.

STG-guided sub-goal generation is triggered only when at least one of the top-3 candidate states has a historical error rate exceeding 0.6, as recorded in the STG. 
Candidate states with no historical records in the STG are skipped. 
When sub-goal generation is triggered, all transitions associated with the candidate states are retrieved from the STG, where each transition is represented as a natural language action accompanied by summarized execution statistics, providing contextual reference for comparison with the current task.
The retrieved transitions are passed to Qwen2.5-VL-7B, which generates a sub-goal to guide the agent's next decision. 
The sub-goal replaces the original task goal in the inference prompt, while the original task goal is retained as additional context. 
Prompt templates are provided in Appendix~\ref{sec:prompt}.

\section{Graph-Structured Analysis of Bottleneck Failures}

Based on the identified bottleneck states and transitions, we further analyze representative failure patterns. 
We manually examine these failures and categorize them into several recurring structural patterns. 
Table~\ref{tab:failure_categories} summarizes the distribution of these failure categories.

\begin{itemize}

\item \textbf{Task Completion Failures (Terminal-State Misidentification).}
These failures occur at terminal or near-terminal states, where agents fail to determine whether the task has been completed correctly. 
In the STG, they manifest as redundant outgoing transitions from terminal states or missing termination edges. 
This suggests that agents lack an explicit notion of task completion and rely on local observations rather than global trajectory context.

\item \textbf{State Perception Failures (State Delta Insensitivity).}
These failures arise when agents fail to perceive state changes after executing an action or incorrectly assume that no further interaction is required. 
In the STG, they appear as repeated transitions between semantically equivalent states or skipped transitions that bypass necessary intermediate states. 
This reflects limited sensitivity to state changes, leading to redundant actions or unstable behavior.

\item \textbf{Instruction Grounding Failures (Constraint Misalignment).}
These failures occur when agents cannot correctly ground multi-constraint instructions into executable steps. 
In the STG, they are characterized by missing or partially executed transitions among multiple outgoing edges that must be completed before reaching the next state. 
This indicates deficiencies in compositional reasoning and constraint integration.

\item \textbf{Interaction Knowledge Failures (Component-Specific Bottlenecks).}
These failures are associated with specific interaction patterns or UI components (e.g., filters or mode switches). 
In the STG, they manifest as consistently high-error transitions concentrated on particular edges shared across trajectories and agents. 
This suggests limited generalization to specialized interaction patterns and insufficient prior knowledge of such components.

\end{itemize}

\begin{table}[t]
    \centering\small
    \renewcommand{\arraystretch}{1.05}
    \setlength{\tabcolsep}{1mm}
    \caption{Distribution of failure categories (\%) observed in GUI agent trajectories. The results reveal a highly structured failure space, where Task Completion Failures (TCF), State Perception Failures (SPF), Instruction Grounding Failures (IGF), and Interaction Knowledge Failures (IKF) account for the majority of errors, while a small portion falls into the Others category.}
    \label{tab:failure_categories}
    \begin{tabular}{l cc cc c}
        \toprule
        Categories & TCF & SPF & IGF & IKF & Others \\ \midrule
        Percentage & 28.7 & 27.3 & 18.2 & 11.2 & 14.6 \\
        \bottomrule
    \end{tabular}
\end{table}

Overall, these findings reveal that failure modes are not only semantically meaningful but also structurally grounded in the transition graph.  
By mapping errors to graph patterns, GUITAR enables a unified diagnosis of GUI agent limitations across perception, reasoning, and interaction.

\section{Prompts}\label{sec:prompt}

\subsection{The Prompts for Building STGs}\label{sec:build_stg_prompt}

\begin{figure*}[htp]
    \centering
    \begin{tcolorbox}[
        colback=mygray, % 主体内容背景颜色
        colframe=darkgray, % 边框颜色
        colbacktitle=darkgray, % 标题背景颜色
        coltitle=white, % 标题文字颜色
        arc=4pt, % 设置圆角大小
        title=Prompt for State Trajectory Abstraction., % 标题内容
        width=\linewidth,
    ]
    \begin{tabular}{p{0.93\linewidth}}
You are an assistant skilled in GUI operations. In a GUI environment, actions are taken according to instructions, and each action transitions from the current page to a new page.

I will provide you with a single task trajectory, including the pages visited during the process and the actions taken. I need you to **abstract this trajectory into a state transition sequence**.

---

**Input Format:**

```
Trajectory: <Goal>

- Page: <Page Description>

  Action: <Action Taken>
  
- Page: <Next Page Description>

  Action: <Action Taken>
  
- ...

- Page: <Final Page Description>

```

---

**Output Format:**

Please provide the abstracted state transition sequence in the following format:

\#\#\# State Transition Sequence

```
<Abstract State 1> --[Abstract Action 1]--> <Abstract State 2> --[Abstract Action 2]--> <Abstract State 3> ...
```

\#\#\# State \& Action Descriptions

| Step | Abstract State | Abstract Action | Next Abstract State |

|------|---------------|-----------------|---------------------|

| 1 | S1: ... | A1: ... | S2: ... |

| 2 | S2: ... | A2: ... | S3: ... |

| ... | | | |

---

**Key Principles:**

- **States** should reflect the **functional role** of the page, not its specific content (e.g., "Product Detail Page" rather than "iPhone 15 Detail Page")

- **Actions** should be described **generically** (e.g., "Select item" rather than "Click iPhone 15")

- Focus on capturing the **essential navigation logic** of the task

    \end{tabular}
    \end{tcolorbox}
    \caption{Prompt for State Trajectory Abstraction.}
    \label{fig:prompt_state_trajectory_abstraction}
\end{figure*}

\begin{figure*}[htp]
    \centering
    \begin{tcolorbox}[
        colback=mygray, % 主体内容背景颜色
        colframe=darkgray, % 边框颜色
        colbacktitle=darkgray, % 标题背景颜色
        coltitle=white, % 标题文字颜色
        arc=4pt, % 设置圆角大小
        title=Prompt for State Transition Graph Construction., % 标题内容
        width=\linewidth,
    ]
    \begin{tabular}{p{0.93\linewidth}}
You are an assistant skilled in GUI operations. In a GUI environment, actions are taken according to instructions, and each action transitions from the current page to a new page. When performing different tasks in an app, similar pages may be encountered; these pages may display different content depending on the task, but the available operations on these pages remain the same.

I will provide you with several different task trajectories, including the pages visited during the process and the actions taken. I need you to:

1. **Analyze** the provided trajectories (page changes and actions between them)

2. **Identify** similar/equivalent pages across different tasks (pages that serve the same functional role even if displaying different content)

3. **Abstract** all trajectories into a unified **state transition graph**, where:

   - **Nodes** = abstract page states (merged similar pages)
   
   - **Edges** = actions that trigger transitions between states

---

**Input Format:**

```
Trajectory Task n: <Goal>
S1: <Page> --> A1: <Action> --> S2: <Page> --> A2: <Action> --> ... --> Sn: <Final Page>
```

---

**Output Format:**

Please provide:

\#\#\# 1. Page Abstraction

List the abstract page states identified, explaining which concrete pages from the trajectories map to each abstract state.

| Abstract State | Concrete Pages (from trajectories) | Description |

|---|---|---|

| S1: ... | Traj1-Page1, Traj2-Page1 | ... |

\#\#\# 2. State Transition Graph

Describe the edges in the format:

```
<Source State> --[Action]--> <Target State>
```

\#\#\# 3. Summary Diagram (Text-based)

Provide a concise visual representation of the full state transition graph.

---

**Key Principles:**

- Pages with the **same functional role** should be merged into one abstract state, even if their content differs

- Actions on edges should be described **generically** (e.g., "click product" rather than "click iPhone 15")

- If a page can be reached via **multiple actions**, represent all corresponding edges

    \end{tabular}
    \end{tcolorbox}
    \caption{Prompt for State Transition Graph Construction.}
    \label{fig:prompt_state_transition_graph_construction}
\end{figure*}

\begin{figure*}[htp]
    \centering
    \begin{tcolorbox}[
        colback=mygray, % 主体内容背景颜色
        colframe=darkgray, % 边框颜色
        colbacktitle=darkgray, % 标题背景颜色
        coltitle=white, % 标题文字颜色
        arc=4pt, % 设置圆角大小
        title=Prompt for State Chain Reconstruction., % 标题内容
        width=\linewidth,
    ]
    \begin{tabular}{p{0.93\linewidth}}
You are an assistant skilled in GUI operations. In GUI environments, actions are performed according to instructions, and each action transitions the interface from the current screen to a new one. When executing different tasks in an application, you often pass through similar screens; these screens may display different content depending on the task, but the available operations are often the same.

We have merged the state transition information from related trajectories into a corresponding state transition diagram. I will provide you with:

1. **The state transition graph** (nodes and edges)

2. **A single task trajectory** to be re-mapped

Your task is to **re-label each state and action** in the trajectory using the abstract nodes and edges from the state transition graph.

---

**Input Format:**

\#\#\# State Transition Graph

**Nodes:**

| Node ID | Description |

|---------|-------------|

| S\_Home | The starting point of the web |

| S\_PregnancyTools | Pages related to pregnancy tools |

| ... | ... |

**Edges:**

```
S\_Home --[Navigate to Pregnancy Section]--> S\_PregnancyTools
S\_PregnancyTools --[Select Calculator]--> S\_CalculatorInput
...
```

---

\#\#\# Trajectory to Re-map

```
S1: <Page Description> --> A1: <Action> --> S2: <Page Description> --> A2: <Action> --> ... --> Sn: <Page Description>
```

---

**Output Format:**

Please output the re-mapped trajectory in the following format:

\#\#\# Re-mapped Trajectory

```
<Node ID> --[<Edge>]--> <Node ID> --[<Edge>]--> ... --> <Node ID>
```

\#\#\# Mapping Details

| Original | Mapped | Reason |

|----------|--------|--------|

| S1: Home Page | S\_Home | Home Page serves as the web entry point |

| A1: Navigate to Pregnancy Section | Navigate to Pregnancy Section | Direct match to graph edge |

| S2: Pregnancy Tools Page | S\_PregnancyTools | Pregnancy Tools Page matches the pregnancy tools node |

| ... | ... | ... |

---

**Key Principles:**

- Map each concrete page to the **most functionally similar** node in the graph

- Map each action to the **closest matching edge** in the graph

- If no exact match exists, choose the **nearest equivalent** and explain your reasoning in the "Reason" column

- Maintain the **sequential order** of the original trajectory

---

    \end{tabular}
    \end{tcolorbox}
    \caption{Prompt for State Chain Reconstruction.}
    \label{fig:prompt_state_chain_reconstruction}
\end{figure*}

\begin{figure*}[t]
    \centering
    \begin{tcolorbox}[
        colback=mygray, % 主体内容背景颜色
        colframe=darkgray, % 边框颜色
        colbacktitle=darkgray, % 标题背景颜色
        coltitle=white, % 标题文字颜色
        arc=4pt, % 设置圆角大小
        title=Prompt for State Assignment. (Part 1), % 标题内容
        width=\linewidth,
    ]
    \begin{tabular}{p{0.93\linewidth}}
You are an assistant skilled in GUI operations. In GUI environments, actions are performed according to instructions, and each action transitions the interface from the current screen to a new one. When executing different tasks in an application, you often pass through similar screens; these screens may display different content depending on the task, but the available operations are often the same.

I will provide you with:

1. **A task trajectory** consisting of a sequence of states and actions

2. **An action history** recording the actions that have been taken so far in the trajectory

3. **A single screen's text description** corresponding to one specific state in the trajectory

4. **The next action** that will be performed on the current screen

Your task is to **map this screen to the most appropriate state** in the state transition trajectory based on its text content and functional role.

---

**Input Format:**

\#\#\# State Description

```

State A: <State description>

... 

```

\#\#\# Task Trajectory

```

S\_A --[Action 1]--> S\_B --[Action 2]--> S\_C --[Action 3]--> ...

```

\#\#\# Action History

```

Step 1: <Action taken>

Step 2: <Action taken>

Step 3: <Action taken>

...

```

\#\#\# Current Action

```

<text content of the screen>

```

\#\#\# Next Execution Action

```

<action that will be performed on the current screen>

```

\#\#\# Screen Description

```

<image of the screen>

```
    \end{tabular}
    \end{tcolorbox}
    \label{fig:prompt_state_assignment}
\end{figure*}

\begin{figure*}[t]
    \centering
    \begin{tcolorbox}[
            colback=mygray,
            colframe=darkgray,
            colbacktitle=darkgray,
            coltitle=white,
            arc=4pt,
            title=Prompt for State Assignment. (Cont.),  % 续接标题
            width=\linewidth,
        ]
        \begin{tabular}{p{0.93\linewidth}}
    ---

**Output Format:**

Please output the result in the following JSON format:

```json

\{

  "mapped\_state": "<State descriptive name>",
  
  "reason": "..."
  
\}

```

---

**Field Descriptions:**

| Field | Type | Description |

|-------|------|-------------|

| `mapped\_state` | `string` | The most appropriate State ID from the graph, or a new descriptive name if no match exists |

| `reason` | `string` | Brief explanation of why this screen was mapped to this state |

---

**Key Principles:**

- Use the **action history** to determine how far along the trajectory the current screen is

- Use the **next execution action** to verify the mapping — the mapped state should have an outgoing edge in the trajectory that matches or is functionally similar to the next action

- Use the **trajectory context** to understand the functional role of the current state (e.g., what actions came before and after)

- Match the screen to the state whose **position in the trajectory** best aligns with the completed actions, current screen content, and next action

- Focus on the **functional purpose** of the screen rather than its specific content details

- Output **only the JSON result**, no additional explanation needed

---
    \end{tabular}
    \end{tcolorbox}
    \caption{Prompt for State Assignment. (Cont.)}
\end{figure*}

\begin{figure*}[htp]
    \centering
    \begin{tcolorbox}[
        colback=mygray, % 主体内容背景颜色
        colframe=darkgray, % 边框颜色
        colbacktitle=darkgray, % 标题背景颜色
        coltitle=white, % 标题文字颜色
        arc=4pt, % 设置圆角大小
        title=Prompt for GUITAR-Guided Inference. (Part 1), % 标题内容
        width=\linewidth,
    ]
    \begin{tabular}{p{0.93\linewidth}}
You are a web operation advisor with access to historical \
execution experiences.

Your job is to analyze the current situation and suggest the most appropriate next action by carefully comparing the current execution context with similar historical experiences.

\#\# Analysis Process (follow this order):

\#\#\# Step 1 - Understand Current Context

- What is the overall task goal?

- What has been done so far (current action history)?

- What does the current screen show?

\#\#\# Step 2 - Task Completion Check (CRITICAL)

Before suggesting any action, explicitly determine whether the task is already complete.

Check the following:

| Checkpoint           | Question to answer                                      |

|----------------------|---------------------------------------------------------|

| Goal achievement     | Has the core objective of the task been fulfilled?      |

| Screen evidence      | Does the current screen visually confirm task success?  |

| History sufficiency  | Have all necessary steps been taken?                    |

| No pending actions   | Is there anything still required to complete the task?  |

Common completion signals to look for on screen:

- Booking/reservation confirmation page is shown

- Search result matches the target criteria

- Form has been submitted successfully

- A confirmation message or success state is visible

Common mistakes that lead to WRONG completion judgment:

- Marking complete when a confirmation click is still needed

- Marking complete when only part of the task is done

- NOT marking complete when the final state is clearly achieved

\#\#\# Step 3 - Compare with Historical Experiences

For each provided historical experience, carefully compare:

- History alignment : How similar is the historical action history to the current one?

  Are they at the same stage? Did they follow the same path?
  
- Screen alignment  : Does the historical screen description match the current screen?

- Error pattern     : What went wrong in that historical state (high error rate)?

- Continuity check  : Is the historical last action similar to the current last action?

\#\#\# Step 4 - Derive Guidance

- Which historical experience is most relevant?

- What mistakes should be avoided?

- What is the most appropriate next action?

    \end{tabular}
    \end{tcolorbox}
    \caption{Prompt for GUITAR-Guided Inference. (Part 1)}
    \label{fig:prompt_guitar_guided_inference}
\end{figure*}

\begin{figure*}[htp]
    \centering
    \begin{tcolorbox}[
        colback=mygray, % 主体内容背景颜色
        colframe=darkgray, % 边框颜色
        colbacktitle=darkgray, % 标题背景颜色
        coltitle=white, % 标题文字颜色
        arc=4pt, % 设置圆角大小
        title=Prompt for GUITAR-Guided Inference. (Cont.), % 标题内容
        width=\linewidth,
    ]
    \begin{tabular}{p{0.93\linewidth}}
\#\# Output Format

Respond in the following JSON format ONLY, no extra text:

\{

    "context\_analysis": \{
    
        "current\_stage": "Describe what stage of the task we are currently at",
        
        "screen\_state": "Describe what the current screen likely shows",
        
        "last\_action\_effect": "What effect did the last action have"
    \},
    
    "completion\_check": \{
    
        "is\_complete": "yes / no",
        
        "evidence": "What on the current screen supports this conclusion?",
        
        "missing": "If not complete, what is still needed? If complete, write N/A"
        
    \},
    
    "experience\_comparison": [
    
        \{
            "experience\_rank": 1,
            
            "history\_alignment": "How well does the historical action history match the current action history?",
            
            "screen\_alignment": "Does the historical screen match the current screen?",
            
            "relevance": "high / medium / low",
            
            "key\_lesson": "The most important lesson learned from this experience"
            
        \}
        
    ],
    
    "next\_step": "COMPLETE if task is done, otherwise a clear description of the next action",
    
    "reason": "Why this action is appropriate",
    
    "caution": "Specific mistakes to avoid"
    
\}

Requirements:

- completion\_check.is\_complete MUST be answered before suggesting any action

- If completion\_check.is\_complete is yes, next\_step MUST be COMPLETE

- experience\_comparison must contain one entry per provided historical experience

- relevance must be one of: high / medium / low

- next\_step must NOT include exact pixel coordinates

- All values must be strings
    \end{tabular}
    \end{tcolorbox}
    \caption{Prompt for GUITAR-Guided Inference. (Cont.)}
\end{figure*}

%%%%%%%%%%%%%%%%%%%%%%%%%%%%%%%%%%%%%%%%%%%%%%%%%%%%%%%%%%%%

% \newpage
% \clearpage

% \input{checklist.tex}

\end{document}